\PassOptionsToPackage{table}{xcolor}
\documentclass[sigconf]{acmart}

\AtBeginDocument{%
  }

\acmConference[]{Preprint}{Under review}{April 2025}

\usepackage{enumitem}
\usepackage{listings}
\usepackage{multirow}
\usepackage{ulem}
\usepackage{float}
\usepackage{tcolorbox}

\definecolor{lightgold}{rgb}{1.0, 0.95, 0.8}

\begin{document}

\settopmatter{printacmref=false}
\renewcommand\footnotetextcopyrightpermission[1]{}

\title[InfiGUI-R1: Advancing Multimodal GUI Agents from Reactive Actors to Deliberative Reasoners]{
\begin{tabular}{@{}l@{\hspace{0.2cm}} c@{}}
\multirow{2}{*}{\includegraphics[width=1.0cm]{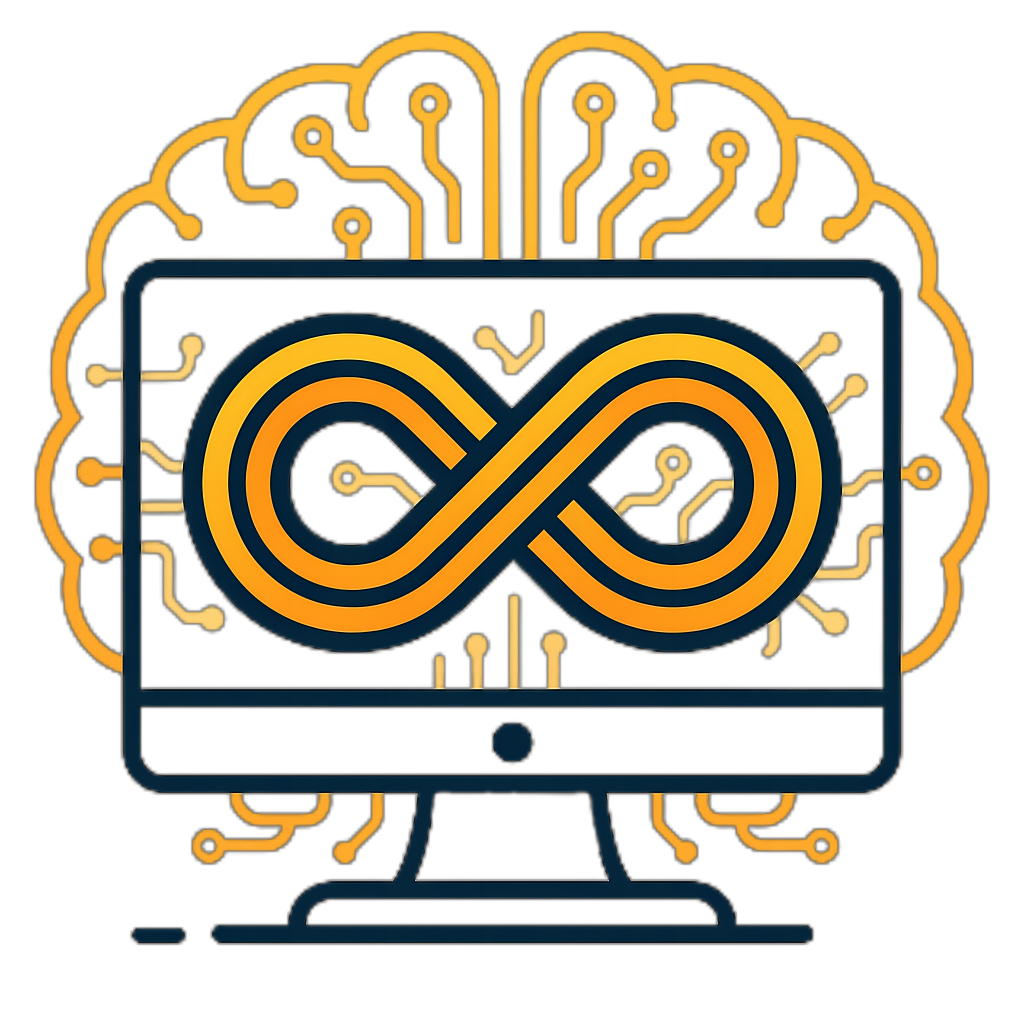}} & InfiGUI-R1: Advancing Multimodal GUI Agents from \\
&Reactive Actors to Deliberative Reasoners
\end{tabular}
}

\author{Yuhang Liu}
\authornote{Both authors contributed equally to this research.}
\affiliation{%
  \institution{Zhejiang University}
}
\email{liuyuhang@zju.edu.cn}

\author{Pengxiang Li}
\authornotemark[1]
\affiliation{%
  \institution{Dalian University of Technology}
}
\email{lipengxiang@mail.dlut.edu.cn}

\author{Congkai Xie}
\affiliation{%
  \institution{Reallm Labs}
}
\email{xieck13@gmail.com}

\author{Xavier Hu}
\affiliation{%
  \institution{Zhejiang University}
}
\email{xavier.hu.research@gmail.com}

\author{Xiaotian Han}
\affiliation{%
}
\email{xiaotian.sky.han@gmail.com}

\author{Shengyu Zhang}
\authornote{Corresponding author.}
\affiliation{%
  \institution{Zhejiang University}
}
\email{sy_zhang@zju.edu.cn}

\author{Hongxia Yang}
\affiliation{%
  \institution{The Hong Kong Polytechnic University}
}
\email{hongxia.yang@polyu.edu.hk}

\author{Fei Wu}
\affiliation{%
  \institution{Zhejiang University}
}
\email{wufei@zju.edu.cn}

\begin{abstract}
Multimodal Large Language Models (MLLMs) have powered Graphical User Interface (GUI) Agents, showing promise in automating tasks on computing devices. 
Recent works have begun exploring reasoning in GUI tasks with encouraging results. However, many current approaches rely on manually designed reasoning templates, which may result in reasoning that is not sufficiently robust and adaptive for complex GUI environments. Meanwhile, some existing agents continue to operate as \textit{Reactive Actors}, relying primarily on implicit reasoning that may lack sufficient depth for GUI tasks demanding planning and error recovery.
We argue that advancing these agents requires a shift from reactive acting towards acting based on deliberate reasoning.
To facilitate this transformation, we introduce \textbf{InfiGUI-R1}, an MLLM-based GUI agent developed through our \textbf{Actor2Reasoner} framework, a reasoning-centric, two-stage training approach designed to progressively evolve agents from \textit{Reactive Actors} to \textit{Deliberative Reasoners}. 
The first stage, \textbf{Reasoning Injection}, focuses on establishing a basic reasoner. We employ \textit{Spatial Reasoning Distillation} to transfer cross-modal spatial reasoning capabilities from teacher models to MLLMs through trajectories with explicit reasoning steps, enabling models to integrate GUI visual-spatial information with logical reasoning before action generation. 
The second stage, \textbf{Deliberation Enhancement}, refines the basic reasoner into a deliberative one using Reinforcement Learning. This stage introduces two approaches: \textit{Sub-goal Guidance}, which rewards models for generating accurate intermediate sub-goals, and \textit{Error Recovery Scenario Construction}, which creates failure-and-recovery training scenarios from identified prone-to-error steps. These approaches enhance the agent's planning abilities and self-correction capabilities.
Experimental results confirm that InfiGUI-R1 achieves strong performance in both cross-platform GUI grounding and trajectory tasks, proving competitive against previous agents, even those with significantly larger parameters.
Resources are available at \url{https://github.com/Reallm-Labs/InfiGUI-R1}.
\end{abstract}


\maketitle

\begin{figure}[th]
    \centering
    \includegraphics[width=\columnwidth]{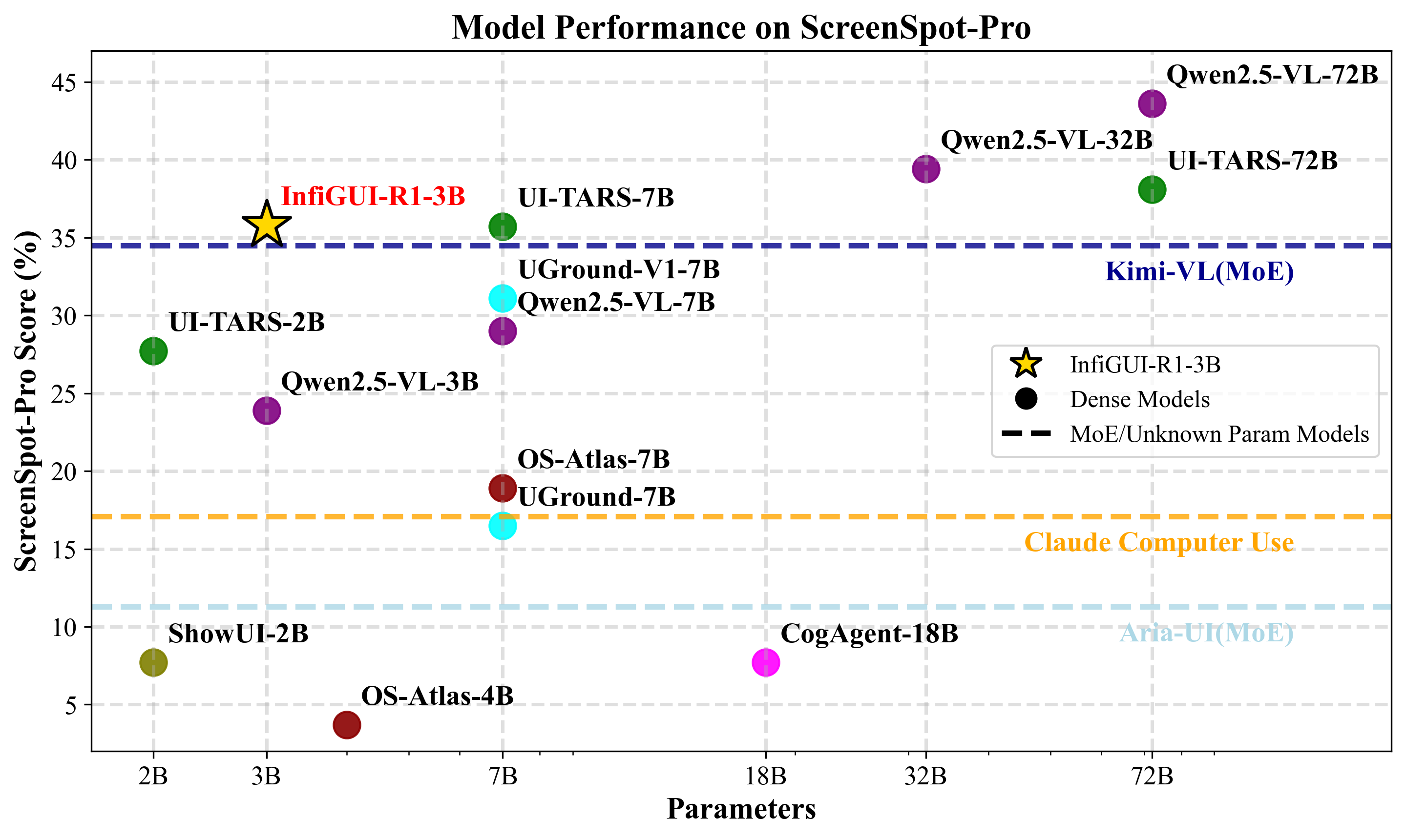}
    \caption{Performance comparison of various GUI agents on the ScreenSpot-Pro benchmark. Our model, InfiGUI-R1-3B marked with a star, demonstrates competitive performance against models with larger parameter counts.}
    \label{fig:screenspot_pro_comparison}
\end{figure}

\section{Introduction}
\label{sec:introduction}

Graphical User Interface (GUI) agents, increasingly powered by advances in Multimodal Large Language Models (MLLMs) \citep{liang2024survey, peng2023kosmos, awadalla2023openflamingo, li2024llava, wang2024qwen2} hold significant promise for automating a wide range of tasks on computing devices such as mobile phones and computers \citep{bonatti2024windows, rawles2024androidworld}. These agents interact with digital environments through visual interfaces, aiming to enhance user productivity and broaden the scope of automated task completion.

Recent works have begun exploring reasoning in GUI tasks with encouraging results. However, many current approaches either rely on manually designed reasoning templates or lack GUI-specific optimization, which may result in reasoning that is not sufficiently robust and adaptive for complex GUI environments. Meanwhile, many existing MLLM-based GUI agents continue to operate as \textit{Reactive Actors} \citep{lin2024showui, cheng2024seeclick}, relying primarily on implicit reasoning. This implicit reasoning often lacks the sufficient depth required for complex, multi-step GUI tasks demanding sophisticated planning and adaptive error recovery. Such tasks necessitate not only precise spatial understanding of dense visual layouts but also the ability to effectively integrate cross-modal information (visual-spatial understanding into textual reasoning) and engage in the deliberative processes crucial for robust, long-horizon task execution.

We argue that fundamentally advancing GUI agent capabilities requires a paradigm shift: moving beyond reactive execution towards agents that function as \textbf{Deliberative Reasoners}. These agents should explicitly incorporate reasoning processes between perception and action (Perception $\rightarrow$ Reasoning $\rightarrow$ Action), enabling them to plan ahead, decompose complex goals, understand spatial relationships deeply, and reflect upon past actions to correct mistakes. This transition is crucial for handling the complexities and dynamic nature of real-world GUI environments.

To enable this transformation, we introduce the \textbf{Actor2Reasoner} framework, a reasoning-centric methodology designed to progressively evolve GUI agents from Reactive Actors to Deliberative Reasoners. Our framework culminates in \textbf{InfiGUI-R1-3B}, an MLLM-based agent demonstrating enhanced reasoning and robustness. The Actor2Reasoner framework tackles two core challenges: 1) reliably injecting foundational reasoning capabilities, particularly bridging the critical cross-modal gap between visual-spatial perception and textual reasoning, to achieve the initial leap from Actor to Reasoner; and 2) refining and elevating the reasoning quality of this foundational Reasoner to instill advanced planning and reflection capabilities, ultimately reaching the Deliberative stage.

The Actor2Reasoner framework unfolds in two distinct stages:

\begin{itemize}[leftmargin=*, nosep]
    \item \textbf{Stage 1: Reasoning Injection (Laying the Foundation for the Reasoner):} This stage focuses on the pivotal transition from Actor to Reasoner. We employ \textbf{Spatial Reasoning Distillation}, leveraging trajectories from a powerful reasoning teacher model that include explicit spatial reasoning steps. By training the base MLLM on this distilled data via Supervised Fine-Tuning (SFT), we guide it to break the direct Perception $\rightarrow$ Action link and explicitly incorporate reasoning, especially spatial reasoning crucial for GUI tasks. This establishes the foundational (Perception $\rightarrow$ Reasoning $\rightarrow$ Action) pattern, overcoming a key limitation of standard MLLMs in integrating visual-spatial understanding into their reasoning flow.

    \item \textbf{Stage 2: Deliberation Enhancement (Refining into a Deliberative Reasoner):} Building upon the Reasoner established in Stage 1, this stage uses Reinforcement Learning (RL) to refine its capabilities towards deliberation. This refinement strategically enhances the two core facets of deliberative reasoning: planning and reflection. Two key innovations drive this process:
        \begin{itemize}[leftmargin=*, nosep, label=\textbullet]
            \item \textbf{Sub-goal Guidance}: To enhance the agent's \textit{forward-looking} planning and task decomposition abilities, we guide it to generate explicit intermediate sub-goals during its reasoning process. The alignment of these generated sub-goals with ground truth provides an intermediate reward signal, effectively training the agent's capacity for proactive planning ("Total Goal $\rightarrow$ Sub-goal $\rightarrow$ Action").
            \item \textbf{Error Recovery Scenario Construction}: Complementing the planning focus, this innovation cultivates the agent's ability to \textit{look backward and adjust} through reflective self-correction – a hallmark of deliberation. We actively construct scenarios within the RL environment that simulate error states or recovery moments (e.g., having just executed an incorrect action or needing to get "back on track" after an error). Training within these scenarios, using targeted rewards, compels the agent to learn adaptive strategies like escaping error states (e.g., using a 'back' action) and adjusting plans after recognizing a mistake. This directly shapes the agent's ability to reflect on its actions and recover from failures, enhancing its robustness.
        \end{itemize}
\end{itemize}

Together, our framework provides a pathway to imbue GUI agents with the reasoning, planning, and reflection capabilities necessary for task automation. We validate the effectiveness of InfiGUI-R1-3B, trained using our Actor2Reasoner framework, on a suite of challenging benchmarks designed to assess core GUI agent competencies. These include tasks requiring precise GUI element grounding across platforms (e.g., ScreenSpot, ScreenSpot-Pro \citep{jurmu2008screenspot, li2025screenspot}) and those demanding complex, long-horizon task execution with planning and adaptation (e.g., AndroidControl\citep{li2024effects}).
Our experimental results demonstrate significant improvements. InfiGUI-R1-3B achieves state-of-the-art cross-platform grounding capabilities (87.5\% avg on ScreenSpot) and strong performance in executing complex, long-horizon tasks (71.1\% success rate on AndroidControl-High) among models with comparable parameter counts.
These findings confirm our framework's ability to cultivate sophisticated planning and reflection abilities, substantially advancing the agent's capacity for deliberate, robust, and effective GUI task automation.

Our main contributions are threefold:
\begin{itemize}[leftmargin=*, nosep]
    \item We propose the \textbf{Actor2Reasoner framework}, a novel two-stage training methodology designed to systematically transform MLLM-based GUI agents from Reactive Actors into Deliberative Reasoners by progressively injecting and refining reasoning capabilities.
    \item We introduce three key technical innovations within this framework: \textbf{Spatial Reasoning Distillation} to establish foundational cross-modal reasoning, \textbf{Sub-goal Guidance} to enhance planning reasoning, and \textbf{Error Recovery Scenario Construction} to cultivate reflective error correction abilities through targeted RL.
    \item We develop \textbf{InfiGUI-R1-3B}, an MLLM-based GUI agent trained via our framework, and demonstrate its effectiveness through comprehensive experiments.
\end{itemize}
\section{Related Works}

\subsection{Multimodal LLMs}
Large Language Models (LLMs) \citep{floridi2020gpt, Touvron2023LLaMAOA, Bai2023QwenTR,xiao2021lawformer} have significantly enhanced the capabilities of AI systems in tackling a wide range of tasks \citep{hu2024infiagentdabench, li2024inficodereval}, thanks to their exceptional ability to process complex semantic and contextual information. The remarkable power of LLMs has also inspired exploration into their potential for processing multimodal data, such as images. Typically, the architecture of Multimodal Large Language Models (MLLMs) consists of three main components: a pre-trained large language model, a trained modality encoder, and a modality interface that connects the LLM with the encoded modality features. Various vision encoders, such as ViT \citep{Dosovitskiy2020AnII}, CLIP \citep{radford2021learning}, and ConvNeXt \citep{liu2022convnet}, extract visual features, which are integrated using techniques like adapter networks \citep{Liu2023VisualIT}, cross-attention layers \citep{Alayrac2022FlamingoAV}, and visual expert modules \citep{wang2023cogvlm}. These methods have facilitated the development of high-performing MLLMs, such as Qwen-VL \citep{Bai2023QwenVLAF}, GPT-4 Vision \citep{openai2023gpt4v}, BLIP-2 \citep{li2023blip} and InfiMM \citep{liu2024infimm}, thus opening new avenues for LLMs in processing GUI tasks.

\subsection{MLLM-based GUI Agents}
Agents are AI systems that perceive their environments, make decisions, and take actions to complete specific tasks. The emergence of LLMs with human-level reasoning ability has significantly advanced the development of agents. For GUI tasks, earlier systems relied on LLMs to read and interpret structured representations such as HTML code \citep{wen2023autodroid}. However, recent works have demonstrated that directly interacting with the visual form of GUIs leads to better performance \citep{202412.2294}. Consequently, MLLM-based GUI agents have been proposed, leveraging visual perception alongside language understanding.

Several representative systems have pioneered this area. ILuvUI \citep{jiang2023iluvui} fine-tuned LLaVA to enhance general GUI comprehension, while AppAgent \citep{zhang2023appagent} explored mobile app usage through autonomous interactions. CogAgent \citep{hong2024cogagent} introduced high-resolution encoders to better capture UI detail, and Ferret-UI-anyres \citep{you2025ferret} supported flexible screen resolutions to handle diverse device settings. 

More recent works have introduced modular and lightweight architectures aimed at improving generalization and deployment efficiency. InfiGUIAgent \citep{liu2025infiguiagent} proposed a two-stage approach, combining general pretraining on grounding and QA tasks with synthetic fine-tuning for hierarchical planning and reasoning. UI-TARS \citep{qin2025ui} extended this by using a unified vision-language interface across mobile, web, and desktop environments, incorporating reflection and milestone tracking mechanisms to boost task success rates. In parallel, AgentS2 \citep{Agent-S2} adopted a generalist-specialist framework, decoupling high-level reasoning from domain-specific grounding modules and enabling long-horizon planning with Mixture of Grounding mechanisms.

In terms of input, recent agents prioritize screenshot-level visual understanding, optionally enhanced with layout or OCR-based textual cues. Techniques such as set-of-mark prompting \citep{yang2023set} and chain-of-action reasoning \citep{pan2024chain} have been employed to improve grounding accuracy and task planning. To further improve interaction efficiency, agents such as UI-R1 \citep{lu2025ui}, GUI-R1 \citep{xia2025gui} replace large-scale supervision with rule-based reinforcement learning, achieving competitive performance with minimal expert data.

Moreover, to support real-world usability, newer agents are tested on increasingly complex environments. UI-TARS and AgentS2 report strong performance on OSWorld and AndroidWorld benchmarks, showing robust cross-platform generalization. GUI-Xplore \citep{sun2025gui} further introduces a one-shot adaptation setting, encouraging agents to build structural UI maps via autonomous exploration before task execution.

\section{Actor2Reasoner}

\begin{figure}[th]
    \centering
    \includegraphics[width=\columnwidth]{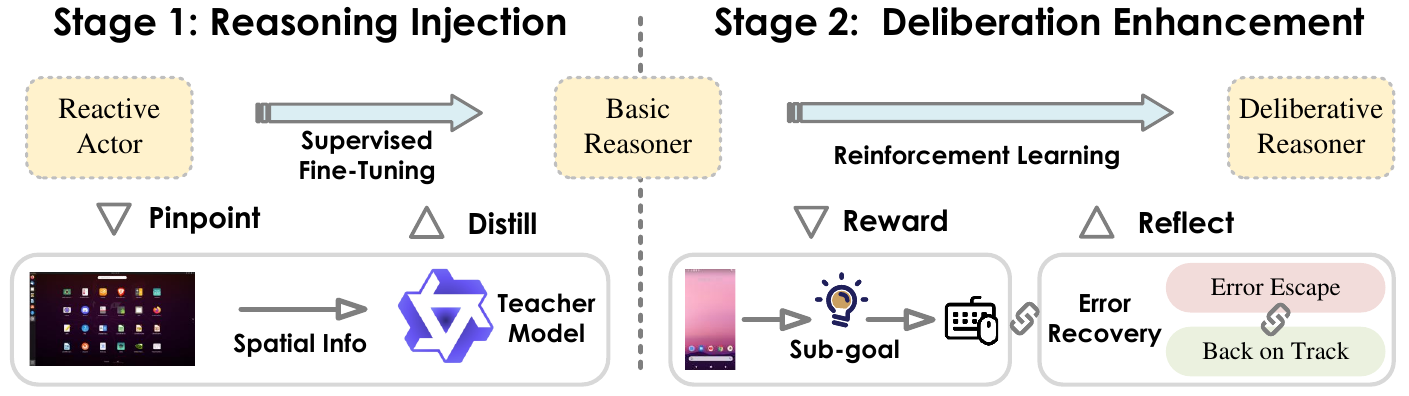}
    \caption{Overview of the \textit{Actor2Reasoner} framework, a two-stage methodology for progressively transforming a Reactive Actor into a Deliberative Reasoner. \textbf{Stage 1: Reasoning Injection} uses Supervised Fine-Tuning (SFT) with Spatial Reasoning Distillation—identifying reasoning bottlenecks (\textit{Pinpoint}) and leveraging a teacher model (\textit{Distill})—to instill foundational cross-modal reasoning and transition the agent into a Basic Reasoner (Perception $\rightarrow$ Reasoning $\rightarrow$ Action). \textbf{Stage 2: Deliberation Enhancement} applies RL to refine planning and reflection capabilities, using Sub-goal Guidance (\textit{Reward}) for forward-looking task decomposition and Error Recovery Scenario Construction (\textit{Reflect}) for backward-looking self-correction, culminating in a Deliberative Reasoner.}
    \label{fig:method}
\end{figure}

We introduce the Actor2Reasoner framework, a reasoning-centric, progressive training methodology designed to systematically enhance the capabilities of Multimodal Large Language Model (MLLM) based GUI agents. The core objective is to transition agents from reactive behavior towards deliberative reasoning for GUI task automation. This framework comprises two stages designed to first establish foundational reasoning and then refine it towards advanced deliberation. Section~\ref{sec:stage1} details the methodology for Stage 1, focusing on reasoning injection via Spatial Reasoning Distillation. Subsequently, Section~\ref{sec:stage2} details the methodology for Stage 2, where RL is employed to enhance deliberation through Sub-goal Guidance and Error Recovery Scenario Construction.

\subsection{Stage 1: Reasoning Injection}
\label{sec:stage1}

The primary objective of Stage 1 is to accomplish the fundamental transition from a \textbf{Reactive Actor} (Perception $\rightarrow$ Action) to a foundational \textbf{Reasoner} (Perception $\rightarrow$ Reasoning $\rightarrow$ Action). This transition is critical because standard MLLMs often struggle to effectively integrate the rich visual-spatial information present in GUI screenshots into their textual reasoning processes. This limitation hinders their ability to handle the GUI tasks that demand precise spatial understanding and grounding. To address this, Stage 1 employs \textbf{Spatial Reasoning Distillation}, which is designed to explicitly inject spatial reasoning capabilities into the agent.

Spatial Reasoning Distillation leverages the reasoning capabilities of a powerful teacher model to generate high-quality reasoning trajectories, which are then used to train the target MLLM (the student). The core idea is to guide the student model to learn not just the correct action, but also the intermediate reasoning steps—particularly those involving spatial logic—that lead to that action. This process is implemented through the following steps:

\subsubsection{Pinpointing Reasoning Bottleneck Samples}

To maximize the efficiency of distillation, we first identify interaction steps where the base MLLM's failure is most likely attributable to a lack of reasoning, rather than fundamental perception or action execution deficits. We term these \textbf{Reasoning Bottleneck Samples}. This identification employs a two-step criterion for each step $s$ in a given trajectory:
\begin{enumerate}[label=(\roman*), nosep, leftmargin=*]
    \item The base MLLM $M$, when given the current GUI screenshot $I_s$ and the overall task goal $G$, fails to predict the correct action. Let $a_\text{high} = M(I_s, G)$.
    \item However, when provided with the additional ground-truth sub-goal $g_s$ for that specific step, the same model $M$ successfully predicts the correct action. Let $a_\text{low} = M(I_s, G, g_s)$.
\end{enumerate}
Formally, the set of reasoning bottleneck steps $S_\text{bottleneck}$ is defined as:
\begin{equation*}
    \begin{split}
        S_\text{bottleneck}& = \\
        \{&s \mid \text{IsCorrect}(a_\text{high}) = \text{False} \land \text{IsCorrect}(a_\text{low}) = \text{True}\}
    \end{split}
\end{equation*}
These samples represent steps where the primary difficulty lies in inferring the immediate task ($g_s$) from the overall goal ($G$) based on the visual context ($I_s$), making them ideal candidates for reasoning injection. We use a base MLLM such as Qwen2.5-VL-3B-Instruct for this filtering process.

\subsubsection{Generating Spatial Reasoning Trajectories}

For each step $s \in S_\text{bottleneck}$, we generate a detailed reasoning trajectory using a high-capability teacher model. This involves:

\paragraph{Spatial Information Extraction and Compression} We extract relevant structural and spatial information (e.g., element types, text content, coordinates, hierarchy) from the accessibility tree (a11y tree) associated with the GUI screenshot $I_s$. Irrelevant attributes and elements are filtered out. A powerful MLLM (e.g., Qwen2.5-VL-32B-Instruct) is then employed to compress this processed information into a concise textual description $D_\text{spatial}$, which consists of a detailed description of the GUI page, including all relevant elements' coordinate information and descriptions for the specific step, capturing the essential spatial layout and key element details.

\paragraph{Reasoning Trajectory Generation} The compressed spatial description $D_\text{spatial}$, the available action space description, and the overall goal $G$ are fed as input to a powerful large language model with strong reasoning capabilities (e.g., QwQ-32B \citep{qwq32b}). This teacher model is prompted to generate both an explicit reasoning text $R_\text{teacher}$ and the corresponding action $a_\text{teacher}$. Crucially, $R_\text{teacher}$ is guided to articulate the logical steps, including using the spatial information in $D_\text{spatial}$ for element localization, relationship assessment, and action justification.

\subsubsection{Injecting Reasoning via SFT}

The generated pairs $(R_\text{teacher}$, $a_\text{teacher})$ are first filtered to ensure quality via rejection sampling based on the correctness of the predicted action $a_\text{teacher}$.
The high-quality pairs are then used to fine-tune the base MLLM. The SFT objective trains the student model to predict the teacher's reasoning and action when given the GUI screenshot and the overall goal: $(I_s, G) \rightarrow (R_\text{teacher}, a_\text{teacher})$. By learning to explicitly generate or implicitly simulate these reasoning steps before outputting the action, the student model internalizes the Perception $\rightarrow$ Reasoning $\rightarrow$ Action pattern.

Upon completion of Stage 1, the resulting model is a foundational \textbf{Reasoner} equipped with enhanced spatial understanding and the basic ability to connect perception to action through an intermediate reasoning process.

\subsection{Stage 2: Deliberation Enhancement}\label{sec:stage2}

Building upon the foundational Reasoner established in Stage 1, the objective of Stage 2 is to refine its capabilities, transforming it into a \textbf{Deliberative Reasoner}. This stage employs RL with rule-based rewards as the primary mechanism for enhancement. The core idea is to cultivate the agent's ability for more sophisticated, "deliberative" decision-making by specifically targeting two aspects: \textit{forward-looking} planning and \textit{backward-looking} reflection/correction. These aspects are addressed through two key innovations integrated into the RL process: \textbf{Sub-goal Guidance} (detailed in Section~\ref{sec:subgoal}) to bolster planning and task decomposition, and \textbf{Error Recovery Scenario Construction} (detailed in Section~\ref{sec:error_recovery}) to foster self-correction and robustness.

\subsubsection{Reinforcement Learning Setup}\label{sec:rl_setup}

We utilize RL to further optimize the agent's policy beyond supervised learning. Specifically, we adopt the \textbf{REINFORCE Leave-One-Out (RLOO)} algorithm \citep{ahmadian2024back}, which effectively reduces the variance of policy gradient estimates by employing the average reward of other samples within the same batch as a baseline for the current sample. This "leave-one-out" baseline strategy obviates the need for training a separate value or critic model, thereby simplifying the training architecture. The RLOO policy gradient $\nabla_\theta J(\theta)$ is estimated as:
\begin{equation*}
    \begin{split}
        \nabla_\theta J(\theta) \approx \\
        \frac{1}{k} \sum_{i=1}^{k} &\left[ R(y_{(i)}, x) - \frac{1}{k-1} \sum_{j \neq i} R(y_{(j)}, x) \right] \nabla_\theta \log \pi_\theta(y_{(i)}|x)
    \end{split}
\end{equation*}
where $k$ is the number of output sequences $y_{(i)}$ sampled from the current policy $\pi_\theta$ given input $x$, and $R(y, x)$ is the reward function evaluating the quality of output $y$.

The design of the reward function $R(y, x)$ is crucial for guiding the agent's learning trajectory. Our total reward $R_\text{total}$ integrates assessments of both output format correctness and task execution accuracy:
\[ R_\text{total} = w_f \cdot R_\text{format} + w_a \cdot R_\text{acc} \]
Here, $R_\text{format}$ checks if the model's output $y$ conforms to the expected format (e.g., putting the reasoning process within \texttt{<think> </think>} tags), yielding 1 if valid and 0 otherwise. $R_\text{acc}$ measures the accuracy of the content, and is calculated \textit{only if} $R_\text{format} = 1$, ensuring the agent first learns to generate structurally valid outputs. $w_f$ and $w_a$ are weighting hyperparameters ($w_f + w_a = 1$).

The accuracy reward $R_\text{acc}$ is tailored to the specific task type:

\paragraph{Agent Trajectory Task Reward ($R_\text{agent}$):} For evaluating sequences of GUI actions, we provide fine-grained feedback by combining rewards for the action type and its parameters:
\[ R_\text{agent} = w_t \cdot R_\text{type} + w_p \cdot R_\text{param} \]
where $w_t + w_p = 1$. $R_\text{type}$ grants a reward of 1 if the predicted action type (e.g., `click`, `type`) matches the ground truth, and 0 otherwise. $R_\text{param}$ provides a stricter reward, granting 1 only if both the action type and \textit{all} its parameters match the ground truth, and 0 otherwise. (Note: This reward is further refined by Sub-goal Guidance in Section~\ref{sec:subgoal}).

\paragraph{Grounding Task Rewards:} For evaluating GUI element localization:
\begin{itemize}[leftmargin=*, nosep]
    \item \textbf{Point Localization Reward ($R_\text{point}$):} Given a predicted point coordinate $(x_p, y_p)$ and the ground-truth bounding box $B_\text{gt}$ of the target element, the reward is 1 if the point falls within the box, and 0 otherwise: 
    \[ R_\text{point} = \mathbb{I}((x_p, y_p) \in B_\text{gt}) \]
    \item \textbf{Bounding Box Reward ($R_\text{bbox}$):} We compute the Intersection over Union (IoU) between the predicted bounding box $B_\text{pred}$ and the ground-truth box $B_\text{gt}$. To avoid penalizing minor deviations excessively while encouraging significant overlap, we use a threshold $\tau_\text{IoU}$. The reward is 1 if the IoU meets or exceeds the threshold, otherwise it is the IoU scaled by the threshold:
    \[ R_\text{bbox} = \begin{cases} 1 & \text{if } \text{IoU}(B_\text{pred}, B_\text{gt}) \ge \tau_\text{IoU} \\ \frac{\text{IoU}(B_\text{pred}, B_\text{gt})}{\tau_\text{IoU}} & \text{if } \text{IoU}(B_\text{pred}, B_\text{gt}) < \tau_\text{IoU} \end{cases} \]
\end{itemize}

\paragraph{Other Task Rewards ($R_\text{other}$):} For auxiliary tasks potentially included in the training mix (e.g., VQA, multiple-choice questions), we use Exact Match (EM) or mathematical expression verification against the ground truth $y_\text{gt}$ to determine correctness:
\[ R_\text{other} = \mathbb{I}(\text{ExactMatch}(y_\text{ans}, y_\text{gt}) \lor \text{MathVerify}(y_\text{ans}, y_\text{gt})) \]

To ensure the agent enhances its GUI-specific deliberation skills without compromising its general multimodal understanding and visual grounding foundations, the RL training phase utilizes a diverse mixture of data. This includes the core GUI trajectory data (e.g., from AndroidControl \citep{li2024effects}), GUI element grounding data (e.g., from widget captioning datasets \citep{li2020widget}), general-purpose multimodal question-answering datasets, and object detection datasets (e.g., from COCO \citep{lin2014microsoft}).

Following established practices for eliciting reasoning \citep{zheng2025easyr1, sheng2024hybridflow}, we employ a system prompt that explicitly instructs the model to first articulate its reasoning process internally before providing the final action. The specific prompt used is:

\begin{tcolorbox}[
    colback=blue!6!white,
    colframe=black,
    colbacktitle=black,
    coltitle=white,
    fonttitle=\bfseries\sffamily,
    title=System Prompt for Reasoning,
    sharp corners,
    boxrule=1pt
]
You FIRST think about the reasoning process as an internal monologue and then provide the final answer.

The reasoning process MUST BE enclosed within <think> </think> tags.
\end{tcolorbox}

\subsubsection{Sub-goal Guidance}
\label{sec:subgoal}

To elevate the foundational Reasoner towards a Deliberative Reasoner capable of sophisticated planning, a core aspect of Stage 2 focuses on enhancing its task decomposition ability. Standard MLLMs often falter when required to independently infer the necessary intermediate steps from a high-level objective within a complex GUI environment. Sub-goal Guidance is specifically designed to address this limitation within the RL framework by incentivizing the agent to formulate and pursue accurate sub-goals, thereby fostering more structured and effective planning. This is achieved by assessing the quality of the sub-goal implied within the agent's reasoning process.

\paragraph{Sub-goal Quality Assessment.}
We incentivize accurate sub-goal formulation by integrating its assessment into the agent's reward structure during RL training. We assess the quality of the implicitly generated sub-goal within the reasoning text.

During training, for each step, we employ a lightweight scoring LLM to analyze the agent's reasoning output (the text within \texttt{<think>...</think>} tags) and attempt to extract the implied sub-goal, denoted as $s_g^\text{extracted}$. This extracted sub-goal $s_g^\text{extracted}$ is then compared against the corresponding ground-truth sub-goal $s_g^\text{gt}$ (obtained from dataset annotations\footnote{\url{https://github.com/google-research/google-research/tree/master/android_control}}). Based on the degree of semantic match between $s_g^\text{extracted}$ and $s_g^\text{gt}$, a raw score $S_\text{raw}$ is assigned on a scale of 1 to 10. If the scoring LLM fails to extract a sub-goal from the reasoning text, $S_\text{raw}$ is set to 0. This raw score is then normalized to the range [0, 1] to produce the final sub-goal reward:
\[ R_\text{subgoal} = \frac{S_\text{raw}}{10} \]
This normalized score, $R_\text{subgoal}$, serves as an intermediate reward signal reflecting the quality of the agent's planning for the current step. To specifically encourage correct planning even when the final action execution fails, we integrate $R_\text{subgoal}$ into the Agent Trajectory Task Reward $R_\text{agent}$ (introduced in Section~\ref{sec:rl_setup}). The formulation is modified as follows, incorporating a dedicated weight $w_s$:
\[ R_\text{agent} = \begin{cases} w_t \cdot R_\text{type} + w_p \cdot R_\text{param} & \text{if } R_\text{param} = 1 \\ w_t \cdot R_\text{type} + w_s \cdot R_\text{subgoal} & \text{if } R_\text{param} = 0 \end{cases} \]
where $w_t, w_p, w_s$ are non-negative weights, and typically $w_s$ is set lower than $w_p$ to prioritize full action correctness when achievable. This conditional reward structure provides targeted feedback on the planning quality when the agent struggles with accurate action execution, thereby guiding the learning process towards better intermediate reasoning and task decomposition.

\begin{table*}[!htp]
    \centering
    \small
    \begin{tabular*}{0.8\textwidth}{@{\extracolsep{\fill}}l ccccccc}
    \toprule
    \multirow{2}{*}{\textbf{Model}} & \multicolumn{6}{c}{\textbf{Accuracy (\%)}} & \multirow{2}{*}{\textbf{Avg.}} \\
    \cmidrule(lr){2-7}
    & \multicolumn{2}{c}{Mobile} & \multicolumn{2}{c}{Desktop} & \multicolumn{2}{c}{Web} & \\
    \cmidrule(lr){2-3} \cmidrule(lr){4-5} \cmidrule(lr){6-7}
    & Text & Icon & Text & Icon & Text & Icon & \\
    \midrule
    \rowcolor{gray!15}
    \multicolumn{8}{l}{\textit{Proprietary Models}} \\
    GPT-4o \citep{gpt4o} & 30.5 & 23.2 & 20.6 & 19.4 & 11.1 & 7.8 & \cellcolor{lightgold}18.8 \\
    Claude Computer Use \citep{anthropic2024b} & - & - & - & - & - & - & \cellcolor{lightgold}83.0 \\
    Gemini 2.0 (Project Mariner) \citep{googldeepmind2024} & - & - & - & - & - & - & \cellcolor{lightgold}84.0 \\
    \midrule
    \rowcolor{gray!15}
    \multicolumn{8}{l}{\textit{General Open-source Models}} \\
    Qwen2-VL-7B \citep{wang2024qwen2} & 61.3 & 39.3 & 52.0 & 45.0 & 33.0 & 21.8 & \cellcolor{lightgold}42.9 \\
    Qwen2.5-VL-3B \citep{bai2025qwen2} & - & - & - & - & - & - & \cellcolor{lightgold}55.5 \\
    Qwen2.5-VL-7B \citep{bai2025qwen2} & - & - & - & - & - & - & \cellcolor{lightgold}\underline{84.7} \\
    \midrule
    \rowcolor{gray!15}
    \multicolumn{8}{l}{\textit{GUI-specific Models}} \\
    CogAgent \citep{hong2024cogagent} & 67.0 & 24.0 & 74.2 & 20.0 & 70.4 & 28.6 & \cellcolor{lightgold}47.4 \\
    SeeClick \citep{cheng2024seeclick} & 78.0 & 52.0 & 72.2 & 30.0 & 55.7 & 32.5 & \cellcolor{lightgold}53.4 \\
    UGround-7B \citep{gou2024navigating} & 82.8 & 60.3 & 82.5 & 63.6 & 80.4 & 70.4 & \cellcolor{lightgold}73.3 \\
    OS-Atlas-7B \citep{wu2024atlas} & 93.0 & 72.9 & 91.8 & 62.9 & 90.9 & 74.3 & \cellcolor{lightgold}82.5 \\
    ShowUI-2B \citep{lin2024showui} & 92.3 & 75.5 & 76.3 & 61.1 & 81.7 & 63.6 & \cellcolor{lightgold}75.1 \\
    Aguvis-7B \citep{huang2024understanding} & \underline{95.6} & 77.7 & \underline{93.8} & 67.1 & 88.3 & 75.2 & \cellcolor{lightgold}84.4 \\
    UI-R1-3B \citep{lu2025ui} & - & - & 90.2 & 59.3 & 85.2 & 73.3 & \cellcolor{lightgold}- \\
    GUI-R1-3B~\citep{xia2025gui} & - & - & \underline{93.8} & 64.8 & 89.6 & 72.1 & \cellcolor{lightgold}- \\
    GUI-R1-7B~\citep{xia2025gui} & - & - & 91.8 & \underline{73.6} & \underline{91.3} & \underline{75.7} & \cellcolor{lightgold}- \\
    UI-TARS-2B \citep{qin2025ui} & 93.0 & 75.5 & 90.7 & 68.6 & 84.3 & 74.8 & \cellcolor{lightgold}82.3 \\
    \midrule
    \rowcolor{gray!15}
    \multicolumn{8}{l}{\textit{Ours}} \\
    \textbf{InfiGUI-R1-3B} & \textbf{97.1} & \textbf{81.2} & \textbf{94.3} & \textbf{77.1} & \textbf{91.7} & \textbf{77.6} & \cellcolor{lightgold}\textbf{87.5} \\
    \bottomrule
    \end{tabular*}
    \caption{Performances on various platforms (Mobile, Desktop, Web) on ScreenSpot. All experiments were conducted using raw screenshot information. Results marked in \textbf{bold} represent the best performance, and those \underline{underlined} indicate the second-best performance.}
    \label{table:screenspot}
\end{table*}

\subsubsection{Error Recovery Scenario Construction}
\label{sec:error_recovery}

While Sub-goal Guidance enhances \textit{forward-looking} planning, developing a \textbf{Deliberative Reasoner} also necessitates the ability to reflect upon and recover from errors—a capability often missing in standard GUI agents prone to irrecoverable failures. To cultivate robustness and adaptability, we utilize \textbf{Error Recovery Scenario Construction}, a technique that directly targets the agent's reflective and corrective reasoning abilities by integrating specific failure-recovery situations into the RL training process. This mechanism complements planning by strengthening the agent's capacity for \textit{backward-looking} adjustment.

\paragraph{Identify Prone-to-error Steps:}
To maximize training efficiency, we first identify interaction steps where the agent demonstrates instability. For a given step $s$, we employ the base model (e.g., Qwen2.5-VL-3B-Instruct) to sample $N_{sample}$ action sequences at a heightened temperature $T$. Steps whose success rate $P_\text{success}(s)$ falls between 0 and 1 ($0 < P_\text{success}(s) < 1$) are designated as \textbf{Prone-to-error Steps}, forming the set $S_\text{error\_prone}$. These steps signify situations where the agent possesses the capacity for correct action but is also susceptible to errors, presenting optimal opportunities for learning corrective strategies. Training on steps the agent always masters or always fails is less efficient for learning recovery; the former needs no correction, while the latter might indicate deeper issues potentially confounded by naive recovery training.

\paragraph{Constructing Recovery Scenarios:}
For each prone-to-error step $s \in S_\text{error\_prone}$, we construct two distinct types of scenarios for RL training, each designed to teach a specific aspect of error handling:

\subparagraph{Error Escape Scenario.} The primary objective here is to train the agent to recognize it has entered an erroneous state and execute an appropriate "escape" action (e.g., pressing the \texttt{back} button). To simulate this, we select an incorrect action $a_s^\text{err}$ sampled during the identification phase, which leads to an unintended subsequent observation $I_{s+1}^\text{err}$. The RL agent is then presented with this error observation $I_{s+1}^\text{err}$ alongside a modified history $H_s^\text{err} = H_{s-1} \oplus a_s^\text{err}$ (where $H_{s-1}$ is the history prior to step $s$, and $\oplus$ denotes concatenation). The desired behavior for the agent in this context is to output a predefined escape action, $a_\text{escape}$.

\subparagraph{Back on Track Scenario.} This scenario aims to train reflective adjustment, enabling the agent to resume the intended task flow after recovering from an error. We assume the agent has just executed the escape action $a_\text{escape}$ from the error state, returning it to the original observation $I_s$ encountered at step $s$. The agent is presented with this original observation $I_s$, but its history reflects the recent detour: $H_s^\text{recover} = H_{s-1} \oplus a_s^\text{err} \oplus a_\text{escape}$. The desired behavior in this "back on track" state is for the agent to perform the originally correct action $a_s^*$ for step $s$, demonstrating its ability to re-evaluate the situation and proceed correctly despite the preceding failure.

The constructed 'Error Escape' and 'Back on Track' scenario samples are incorporated into the data used for RL training in Stage 2. When the agent encounters these scenarios as input $x$ and generates an output $y$, its performance is evaluated using the same comprehensive reward function $R_\text{total}(y, x)$. By rewarding successful escape actions in the first scenario type and correct subsequent actions in the second, the RL process specifically reinforces the agent's adaptive strategies for handling failures. This targeted training solidifies the agent's transition towards a Deliberative Reasoner, together with the task decomposition ability.
\section{Experiments}
\begin{table*}[!htp]
    \centering
    \small
    \setlength{\tabcolsep}{0pt}
    \begin{tabular*}{\textwidth}{@{\extracolsep{\fill}}l *{21}{c}}
    \toprule
    \multirow{3}{*}{\textbf{Model}} & \multicolumn{3}{c}{\textbf{CAD}} & \multicolumn{3}{c}{\textbf{Development}} & \multicolumn{3}{c}{\textbf{Creative}} & \multicolumn{3}{c}{\textbf{Scientific}} & \multicolumn{3}{c}{\textbf{Office}} & \multicolumn{3}{c}{\textbf{OS}} & \multicolumn{3}{c}{\textbf{Avg.}} \\
    \cmidrule(lr){2-4} \cmidrule(lr){5-7} \cmidrule(lr){8-10} \cmidrule(lr){11-13} \cmidrule(lr){14-16} \cmidrule(lr){17-19} \cmidrule(lr){20-22}
    & Text & Icon & Avg. & Text & Icon & Avg. & Text & Icon & Avg. & Text & Icon & Avg. & Text & Icon & Avg. & Text & Icon & Avg. & Text & Icon & Avg. \\
    \midrule
    \rowcolor{gray!15}
    \multicolumn{22}{l}{\textit{Proprietary Models}} \\
    GPT-4o~\citep{hurst2024gpt} & 2.0 & 0.0 & 1.5 & 1.3 & 0.0 & 0.7 & 1.0 & 0.0 & 0.6 & 2.1 & 0.0 & 1.2 & 1.1 & 0.0 & 0.9 & 0.0 & 0.0 & 0.0 & 1.3 & 0.0 & \cellcolor{lightgold}0.8 \\
    Claude Computer Use~\citep{anthropic2024b} & 14.5 & 3.7 & 11.9 & 22.0 & 3.9 & 12.6 & 25.9 & 3.4 & 16.8 & 33.9 & 15.8 & 25.8 & 30.1 & 16.3 & 26.9 & 11.0 & 4.5 & 8.1 & 23.4 & 7.1 & \cellcolor{lightgold}17.1 \\
    \midrule
    \rowcolor{gray!15}
    \multicolumn{22}{l}{\textit{General Open-source Models}} \\
    Qwen2-VL-7B~\citep{wang2024qwen2} & 0.5 & 0.0 & 0.4 & 2.6 & 0.0 & 1.3 & 1.5 & 0.0 & 0.9 & 6.3 & 0.0 & 3.5 & 3.4 & 1.9 & 3.0 & 0.9 & 0.0 & 0.5 & 2.5 & 0.2 & \cellcolor{lightgold}1.6 \\
    Qwen2.5-VL-3B~\citep{bai2025qwen2} & - & - & - & - & - & - & - & - & - & - & - & - & - & - & - & - & - & - & - & - & \cellcolor{lightgold}23.9 \\
    Qwen2.5-VL-7B~\citep{bai2025qwen2} & - & - & - & - & - & - & - & - & - & - & - & - & - & - & - & - & - & - & - & - & \cellcolor{lightgold}29.0 \\
    Kimi-VL~\citep{team2025kimi} & - & - & - & - & - & - & - & - & - & - & - & - & - & - & - & - & - & - & - & - & \cellcolor{lightgold}\underline{34.5} \\
    \midrule
    \rowcolor{gray!15}
    \multicolumn{22}{l}{\textit{GUI-specific Models}} \\
    SeeClick~\citep{cheng2024seeclick} & 2.5 & 0.0 & 1.9 & 0.6 & 0.0 & 0.3 & 1.0 & 0.0 & 0.6 & 3.5 & 0.0 & 2.0 & 1.1 & 0.0 & 0.9 & 2.8 & 0.0 & 1.5 & 1.8 & 0.0 & \cellcolor{lightgold}1.1 \\
    CogAgent-18B~\citep{hong2024cogagent} & 7.1 & 3.1 & 6.1 & 14.9 & 0.7 & 8.0 & 9.6 & 0.0 & 5.6 & 22.2 & 1.8 & 13.4 & 13.0 & 0.0 & 10.0 & 5.6 & 0.0 & 3.1 & 12.0 & 0.8 & \cellcolor{lightgold}7.7 \\
    Aria-UI~\citep{yang2024aria} & 7.6 & 1.6 & 6.1 & 16.2 & 0.0 & 8.4 & 23.7 & 2.1 & 14.7 & 27.1 & 6.4 & 18.1 & 20.3 & 1.9 & 16.1 & 4.7 & 0.0 & 2.6 & 17.1 & 2.0 & \cellcolor{lightgold}11.3 \\
    OS-Atlas-4B~\citep{wu2024atlas} & 2.0 & 0.0 & 1.5 & 7.1 & 0.0 & 3.7 & 3.0 & 1.4 & 2.3 & 9.0 & 5.5 & 7.5 & 5.1 & 3.8 & 4.8 & 5.6 & 0.0 & 3.1 & 5.0 & 1.7 & \cellcolor{lightgold}3.7 \\
    OS-Atlas-7B~\citep{wu2024atlas} & 12.2 & 4.7 & 10.3 & 33.1 & 1.4 & 17.7 & 28.8 & 2.8 & 17.9 & 37.5 & 7.3 & 24.4 & 33.9 & 5.7 & 27.4 & 27.1 & 4.5 & 16.8 & 28.1 & 4.0 & \cellcolor{lightgold}18.9 \\
    ShowUI-2B~\citep{lin2024showui} & 2.5 & 0.0 & 1.9 & 16.9 & 1.4 & 9.4 & 9.1 & 0.0 & 5.3 & 13.2 & 7.3 & 10.6 & 15.3 & 7.5 & 13.5 & 10.3 & 2.2 & 6.6 & 10.8 & 2.6 & \cellcolor{lightgold}7.7 \\
    UGround-7B \citep{gou2024navigating} & 14.2 & 1.6 & 11.1 & 26.6 & 2.1 & 14.7 & 27.3 & 2.8 & 17.0 & 31.9 & 2.7 & 19.3 & 31.6 & 11.3 & 27.0 & 17.8 & 0.0 & 9.7 & 25.0 & 2.8 & \cellcolor{lightgold}16.5 \\
    UGround-V1-7B~\citep{gou2024navigating} & - & - & 13.5 & - & - & \underline{35.5} & - & - & 27.8 & - & - & 38.8 & - & - & 48.8 & - & - & \underline{26.1} & - & - & \cellcolor{lightgold}31.1 \\
    UI-R1-3B~\citep{lu2025ui} & 11.2 & 6.3 & - & 22.7 & 4.1 & - & 27.3 & 3.5 & - & 42.4 & 11.8 & - & 32.2 & 11.3 & - & 13.1 & 4.5  & - & - & - & \cellcolor{lightgold}17.8 \\
    GUI-R1-3B~\citep{xia2025gui} & \underline{26.4} & 7.8 & - & 33.8 & \underline{4.8} & - & 40.9 & 5.6 & - & \underline{61.8} & 17.3 & - & 53.6 & 17.0 & - & 28.1 & 5.6 & - & - & - & \cellcolor{lightgold}- \\
    GUI-R1-7B~\citep{xia2025gui} & 23.9 & 6.3 & - & 49.4 & \underline{4.8} & - & 38.9 & \underline{8.4} & - & 55.6 & 11.8 & - & 58.7 & \underline{26.4} & - & \underline{42.1} & \textbf{16.9} & - & - & - & \cellcolor{lightgold}- \\
    UI-TARS-2B~\citep{qin2025ui} & 17.8 & 4.7 & 14.6 & 47.4 & 4.1 & 26.4 & 42.9 & 6.3 & 27.6 & 56.9 & 17.3 & 39.8 & 50.3 & 17.0 & 42.6 & 21.5 & 5.6 & 14.3 & 39.6 & 8.4 & \cellcolor{lightgold}27.7 \\
    UI-TARS-7B~\citep{qin2025ui} & 20.8 & \underline{9.4} & \underline{18.0} & \textbf{58.4} & \textbf{12.4} & \textbf{36.1} & \textbf{50.0} & \textbf{9.1} & \textbf{32.8} & \textbf{63.9} & \textbf{31.8} & \textbf{50.0} & \underline{63.3} & 20.8 & \underline{53.5} & 30.8 & \textbf{16.9} & 24.5 & \underline{47.8} & \textbf{16.2} & \cellcolor{lightgold}\textbf{35.7} \\
    \midrule
    \rowcolor{gray!15}
    \multicolumn{22}{l}{\textit{Ours}} \\
     \textbf{InfiGUI-R1-3B} & \textbf{33.0} & \textbf{14.1} & \textbf{28.4} & \underline{51.3} & \textbf{12.4} & 32.4 & \underline{44.9} & 7.0 & \underline{29.0} & 58.3 & \underline{20.0} & \underline{41.7} & \textbf{65.5} & \textbf{28.3} & \textbf{57.0} & \textbf{43.9} & \underline{12.4} & \textbf{29.6} & \textbf{49.1} & \underline{14.1} & \cellcolor{lightgold}\textbf{35.7} \\
    \bottomrule
    \end{tabular*}
    \caption{Performance comparison of different agent models across various task categories based on Text, Icon, and Average scores on ScreenSpot-Pro. Results marked in \textbf{bold} represent the best performance, and those \underline{underlined} indicate the second-best performance.}
    \label{tab:screenspot_pro}
\end{table*}

In this section, we detail the experimental setup used to train and evaluate our proposed \textbf{InfiGUI-R1-3B} agent. We describe the implementation details, the benchmarks used for evaluation, and present a comprehensive analysis of the results compared to existing state-of-the-art methods.

\subsection{Setup}

\paragraph{Implementation Details.} Our model, \textbf{InfiGUI-R1-3B}, is built upon Qwen2.5-VL-3B-Instruct and trained using the proposed \textbf{Actor2Reasoner} Framework, which consists of two main stages.
For the RL reward function $R_\text{total} = w_f \cdot R_\text{format} + w_a \cdot R_\text{acc}$, we set the weights $w_f = 0.1$ and $w_a = 0.9$. Within the agent trajectory accuracy reward $R_\text{acc\_agent}$, the weights are $w_t = 0.2$ for type matching and $w_p = 0.8$ for exact parameter matching. For bounding box rewards ($R_\text{bbox}$), the IoU threshold is $\tau_\text{IoU} = 0.7$. When using sub-goal similarity as a reward ($R_\text{subgoal}$) in cases where the action parameters are incorrect ($R_\text{param}=0$), we use a weight $w_s = 0.2$.

\paragraph{Training Data.} To ensure both strong GUI capabilities and general multimodal understanding, we train \textbf{InfiGUI-R1-3B} on a diverse dataset mixture: AndroidControl (10k trajectories + 2k reflection-focused trajectories), GUI Grounding data (5k samples aggregated from RicoSCA, Widget Caption, etc.), MathV360K (11k samples for general reasoning), and COCO (4k samples for general visual grounding and understanding).

\paragraph{Training Parameters.} All experiments were conducted using 16 NVIDIA H800 GPUs. For the SFT stage (Stage 1), we used a learning rate of 2.0e-6, a global batch size of 32, and a warmup ratio of 0.1. For the RL stage (Stage 2), we used a learning rate of 1.0e-6, a batch size of 256 for training updates, a rollout batch size of 256, and generated 16 rollouts per sample during policy exploration.

\subsection{Evaluation Benchmarks}

To comprehensively evaluate \textbf{InfiGUI-R1-3B}, we utilize several key benchmarks targeting different facets of GUI agent capabilities:

\paragraph{ScreenSpot \& ScreenSpot-Pro:} These benchmarks assess fundamental GUI understanding and element grounding accuracy across diverse platforms (Mobile, Desktop, Web). ScreenSpot-Pro specifically increases the difficulty with complex desktop applications and high-resolution screens.

\paragraph{AndroidControl:} This benchmark evaluate the agent's ability to execute complex, multi-step tasks within realistic Android environments. They directly test the higher-level reasoning capabilities crucial for a Deliberative Reasoner, including planning, and state tracking over long interaction trajectories. We report results on the Low-level (Low) and High-level (High) splits of AndroidControl.

\subsection{Results}

We compare \textbf{InfiGUI-R1-3B} against a range of state-of-the-art open-source and proprietary GUI agents. The results demonstrate the effectiveness of our Actor2Reasoner framework in advancing GUI agents towards deliberative reasoning.

\begin{table}[ht]
    \centering
    \small
    \setlength{\tabcolsep}{3pt}
    \begin{tabular*}{\columnwidth}{@{\extracolsep{\fill}}l ccc ccc}
    \toprule
    \multirow{2}{*}{\textbf{Model}} & \multicolumn{3}{c}{\textbf{AndroidControl-Low}} & \multicolumn{3}{c}{\textbf{AndroidControl-High}} \\
    \cmidrule(lr){2-4} \cmidrule(lr){5-7}
    & Type & Grounding & SR & Type & Grounding & SR \\
    \midrule
    Claude* & 74.3 & 0.0 & \cellcolor{lightgold}19.4 & 63.7 & 0.0 & \cellcolor{lightgold}12.5 \\
    GPT-4o & 74.3 & 0.0 & \cellcolor{lightgold}19.4 & 66.3 & 0.0 & \cellcolor{lightgold}20.8 \\
    Aria-UI & – & \underline{87.7} & \cellcolor{lightgold}67.3 & – & 43.2 & \cellcolor{lightgold}10.2 \\
    OS-Atlas-4B & 91.9 & 83.8 & \cellcolor{lightgold}80.6 & \textbf{84.7} & 73.8 & \cellcolor{lightgold}67.5 \\
    Aguvis-7B & – & – & \cellcolor{lightgold}80.5 & – & – & \cellcolor{lightgold}61.5 \\
    Aguvis-72B & – & – & \cellcolor{lightgold}84.4 & – & – & \cellcolor{lightgold}66.4 \\
    UI-R1 & 94.3 & 82.6 & \cellcolor{lightgold}- & - & - & \cellcolor{lightgold}- \\
    GUI-R1-3B & - & - & \cellcolor{lightgold}- & 58.0 & 56.2 & \cellcolor{lightgold}46.6\\
    GUI-R1-7B & - & - & \cellcolor{lightgold}- & 71.6 & 65.6 & \cellcolor{lightgold}51.7\\
    UI-TARS-2B & \textbf{98.1} & 87.3 & \cellcolor{lightgold}\underline{89.3} & 81.2 & \textbf{78.4} & \cellcolor{lightgold}\underline{68.9} \\
    \midrule
    \rowcolor{gray!15}
    \multicolumn{7}{l}{\textit{Ours}} \\
    InfiGUI-R1-3B & \underline{96.0} & \textbf{93.2} & \cellcolor{lightgold}\textbf{92.1} & \underline{82.7} & \underline{74.4} & \cellcolor{lightgold}\textbf{71.1} \\
    \bottomrule
    \end{tabular*}
    \caption{Performance comparison of different agent models on AndroidControl benchmarks. SR stands for Success Rate. Results marked in \textbf{bold} represent the best performance, and those \underline{underlined} indicate the second-best performance.}
    \label{table:androidcontrol}
\end{table}

\paragraph{Performance on ScreenSpot.} Table~\ref{table:screenspot} summarizes the results on the ScreenSpot benchmark, evaluating grounding across Mobile, Desktop, and Web platforms. \textbf{InfiGUI-R1-3B} achieves state-of-the-art performance among all compared models, including proprietary ones like Gemini 1.5 Pro and Claude, with an impressive average accuracy of \textbf{87.5\%}. It consistently ranks first across all platforms and both text-based and icon-based grounding tasks (Mobile: 97.1/81.2, Desktop: 94.3/77.1, Web: 91.7/77.6). This outstanding performance underscores the robustness and generalization ability of \textbf{InfiGUI-R1-3B}'s visual understanding and grounding capabilities.

\paragraph{Performance on ScreenSpot-Pro.} As shown in Table~\ref{tab:screenspot_pro}, \textbf{InfiGUI-R1-3B} achieves competitive performance on the demanding Screen-Spot-Pro benchmark, which focuses on complex, high-resolution desktop GUI grounding.
\textcolor{red!60!black}{\uline{With an overall average score of 35.7, it performs comparably to the larger UI-TARS-7B model (35.7) and significantly outperforms other baselines like OS-Atlas-7B (18.9) and UGround-7B (16.5).}}
Our model shows particular strength in categories like CAD (28.4 avg), Office (57.0 avg) and OS (29.6 avg), demonstrating robust grounding capabilities even in specialized software environments. While not universally outperforming the top model in every category, the strong overall performance validates the effectiveness of our approach.

\paragraph{Performance on AndroidControl.} Table~\ref{table:androidcontrol} presents the results on the AndroidControl benchmark. \textbf{InfiGUI-R1-3B} achieves a high Success Rate (SR) of 92.1\% on AndroidControl-Low and 71.1\% on AndroidControl-High.
\textcolor{red!60!black}{\uline{This surpasses the previous state-of-the-art model with similar parameters, UI-TARS-2B (SR: 89.3\% / 68.9\%). Furthermore, it also outperforms larger GUI-specific models such as Aguvis-72B (SR: 84.4\% / 66.4\%).}}
This highlights the effectiveness of the training focused on planning capabilities in our Stage 2.

In summary, the experimental results across AndroidControl, ScreenSpot-Pro, and ScreenSpot demonstrate that \textbf{InfiGUI-R1-3B} significantly advances the capabilities of GUI agents. Our Actor2Reasoner framework, combining Spatial Reasoning Distillation and RL-based Deliberation Enhancement (Sub-goal Guidance, Error Recovery), successfully transforms a base MLLM into a more effective Deliberative Reasoner,
achieving state-of-the-art among models with similar parameter counts in trajectory-based tasks and element grounding across different platforms and resolutions, even with a relatively small 3B parameter model.

\subsection{Visualization}
\begin{figure}[th]
    \centering
    \includegraphics[width=\columnwidth]{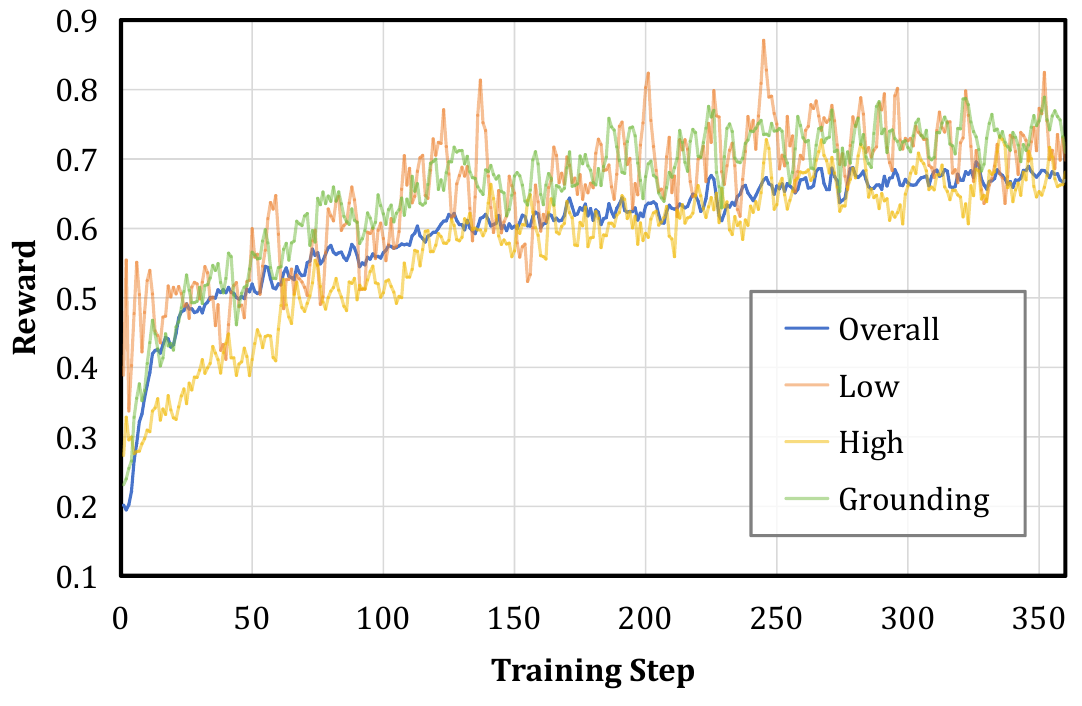}
    \caption{Reward curves during reinforcement learning training. The plot shows the overall reward and the rewards for individual task types (Low-level, High-level, Grounding) over training steps.}
    \label{fig:reward_curve}
\end{figure}

Figure \ref{fig:reward_curve} illustrates the reward progression throughout the reinforcement learning training process. It displays both the overall reward accumulated by the agent and the specific rewards obtained for different task categories - Low-level, High-level, and Grounding tasks. As observed from the curves, the rewards for the overall agent performance, as well as for each individual task type, exhibit a consistent upward trend as training progresses. This indicates that the agent effectively learns and improves its performance across all GUI tasks during the RL training phase.
\section{Conclusion}
We present \textbf{InfiGUI-R1-3B}, a multimodal GUI agent that bridges the gap between reactive execution and deliberative reasoning. Through the Actor2Reasoner framework, our approach systematically injects and refines reasoning capabilities in two stages: \textit{Spatial Reasoning Distillation} to build foundational cross-modal reasoning, and \textit{Deliberation Enhancement} via reinforcement learning to support sub-goal planning and error recovery. Empirical results across diverse benchmarks demonstrate that InfiGUI-R1-3B not only matches or surpasses larger models in grounding accuracy but also excels in long-horizon task execution with robust planning and reflection.

\bibliographystyle{ACM-Reference-Format}
\bibliography{bibliography}


\begin{thebibliography}{58}


\ifx \showCODEN    \undefined \def \showCODEN     #1{\unskip}     \fi
\ifx \showISBNx    \undefined \def \showISBNx     #1{\unskip}     \fi
\ifx \showISBNxiii \undefined \def \showISBNxiii  #1{\unskip}     \fi
\ifx \showISSN     \undefined \def \showISSN      #1{\unskip}     \fi
\ifx \showLCCN     \undefined \def \showLCCN      #1{\unskip}     \fi
\ifx \shownote     \undefined \def \shownote      #1{#1}          \fi
\ifx \showarticletitle \undefined \def \showarticletitle #1{#1}   \fi
\ifx \showURL      \undefined \def \showURL       {\relax}        \fi
\providecommand\bibfield[2]{#2}
\providecommand\bibinfo[2]{#2}
\providecommand\natexlab[1]{#1}
\providecommand\showeprint[2][]{arXiv:#2}

\bibitem[Agashe et~al\mbox{.}(2025)]%
        {Agent-S2}
\bibfield{author}{\bibinfo{person}{Saaket Agashe}, \bibinfo{person}{Kyle Wong}, \bibinfo{person}{Vincent Tu}, \bibinfo{person}{Jiachen Yang}, \bibinfo{person}{Ang Li}, {and} \bibinfo{person}{Xin~Eric Wang}.} \bibinfo{year}{2025}\natexlab{}.
\newblock \bibinfo{title}{Agent S2: A Compositional Generalist-Specialist Framework for Computer Use Agents}.
\newblock
\showeprint[arxiv]{2504.00906}~[cs.AI]
\urldef\tempurl%
\url{https://arxiv.org/abs/2504.00906}
\showURL{%
\tempurl}


\bibitem[Ahmadian et~al\mbox{.}(2024)]%
        {ahmadian2024back}
\bibfield{author}{\bibinfo{person}{Arash Ahmadian}, \bibinfo{person}{Chris Cremer}, \bibinfo{person}{Matthias Gall{\'e}}, \bibinfo{person}{Marzieh Fadaee}, \bibinfo{person}{Julia Kreutzer}, \bibinfo{person}{Olivier Pietquin}, \bibinfo{person}{Ahmet {\"U}st{\"u}n}, {and} \bibinfo{person}{Sara Hooker}.} \bibinfo{year}{2024}\natexlab{}.
\newblock \showarticletitle{Back to basics: Revisiting reinforce style optimization for learning from human feedback in llms}.
\newblock \bibinfo{journal}{\emph{arXiv preprint arXiv:2402.14740}} (\bibinfo{year}{2024}).
\newblock


\bibitem[Alayrac et~al\mbox{.}(2022)]%
        {Alayrac2022FlamingoAV}
\bibfield{author}{\bibinfo{person}{Jean-Baptiste Alayrac}, \bibinfo{person}{Jeff Donahue}, \bibinfo{person}{Pauline Luc}, \bibinfo{person}{Antoine Miech}, \bibinfo{person}{Iain Barr}, \bibinfo{person}{Yana Hasson}, \bibinfo{person}{Karel Lenc}, \bibinfo{person}{Arthur Mensch}, \bibinfo{person}{Katie Millican}, \bibinfo{person}{Malcolm Reynolds}, \bibinfo{person}{Roman Ring}, \bibinfo{person}{Eliza Rutherford}, \bibinfo{person}{Serkan Cabi}, \bibinfo{person}{Tengda Han}, \bibinfo{person}{Zhitao Gong}, \bibinfo{person}{Sina Samangooei}, \bibinfo{person}{Marianne Monteiro}, \bibinfo{person}{Jacob Menick}, \bibinfo{person}{Sebastian Borgeaud}, \bibinfo{person}{Andy Brock}, \bibinfo{person}{Aida Nematzadeh}, \bibinfo{person}{Sahand Sharifzadeh}, \bibinfo{person}{Mikolaj Binkowski}, \bibinfo{person}{Ricardo Barreira}, \bibinfo{person}{Oriol Vinyals}, \bibinfo{person}{Andrew Zisserman}, {and} \bibinfo{person}{Karen Simonyan}.} \bibinfo{year}{2022}\natexlab{}.
\newblock \showarticletitle{Flamingo: a Visual Language Model for Few-Shot Learning}. In \bibinfo{booktitle}{\emph{Advances in Neural Information Processing Systems 35: Annual Conference on Neural Information Processing Systems 2022, NeurIPS 2022, New Orleans, LA, USA, November 28 - December 9, 2022}}, \bibfield{editor}{\bibinfo{person}{Sanmi Koyejo}, \bibinfo{person}{S.~Mohamed}, \bibinfo{person}{A.~Agarwal}, \bibinfo{person}{Danielle Belgrave}, \bibinfo{person}{K.~Cho}, {and} \bibinfo{person}{A.~Oh}} (Eds.).
\newblock
\urldef\tempurl%
\url{http://papers.nips.cc/paper\_files/paper/2022/hash/960a172bc7fbf0177ccccbb411a7d800-Abstract-Conference.html}
\showURL{%
\tempurl}


\bibitem[Anthropic(2024)]%
        {anthropic2024b}
\bibfield{author}{\bibinfo{person}{Anthropic}.} \bibinfo{year}{2024}\natexlab{}.
\newblock \bibinfo{title}{Developing a computer use model}.
\newblock \bibinfo{howpublished}{\url{https://www.anthropic.com/news/developing-computer-use}}.
\newblock
\newblock
\shownote{Accessed: 2025-04-12}.


\bibitem[Awadalla et~al\mbox{.}(2023)]%
        {awadalla2023openflamingo}
\bibfield{author}{\bibinfo{person}{Anas Awadalla}, \bibinfo{person}{Irena Gao}, \bibinfo{person}{Josh Gardner}, \bibinfo{person}{Jack Hessel}, \bibinfo{person}{Yusuf Hanafy}, \bibinfo{person}{Wanrong Zhu}, \bibinfo{person}{Kalyani Marathe}, \bibinfo{person}{Yonatan Bitton}, \bibinfo{person}{Samir Gadre}, \bibinfo{person}{Shiori Sagawa}, {et~al\mbox{.}}} \bibinfo{year}{2023}\natexlab{}.
\newblock \showarticletitle{Openflamingo: An open-source framework for training large autoregressive vision-language models}.
\newblock \bibinfo{journal}{\emph{arXiv preprint arXiv:2308.01390}} (\bibinfo{year}{2023}).
\newblock


\bibitem[Bai et~al\mbox{.}(2023a)]%
        {Bai2023QwenTR}
\bibfield{author}{\bibinfo{person}{Jinze Bai}, \bibinfo{person}{Shuai Bai}, \bibinfo{person}{Yunfei Chu}, \bibinfo{person}{Zeyu Cui}, \bibinfo{person}{Kai Dang}, \bibinfo{person}{Xiaodong Deng}, \bibinfo{person}{Yang Fan}, \bibinfo{person}{Wenhang Ge}, \bibinfo{person}{Yu Han}, \bibinfo{person}{Fei Huang}, \bibinfo{person}{Binyuan Hui}, \bibinfo{person}{Luo Ji}, \bibinfo{person}{Mei Li}, \bibinfo{person}{Junyang Lin}, \bibinfo{person}{Runji Lin}, \bibinfo{person}{Dayiheng Liu}, \bibinfo{person}{Gao Liu}, \bibinfo{person}{Chengqiang Lu}, \bibinfo{person}{K. Lu}, \bibinfo{person}{Jianxin Ma}, \bibinfo{person}{Rui Men}, \bibinfo{person}{Xingzhang Ren}, \bibinfo{person}{Xuancheng Ren}, \bibinfo{person}{Chuanqi Tan}, \bibinfo{person}{Sinan Tan}, \bibinfo{person}{Jianhong Tu}, \bibinfo{person}{Peng Wang}, \bibinfo{person}{Shijie Wang}, \bibinfo{person}{Wei Wang}, \bibinfo{person}{Shengguang Wu}, \bibinfo{person}{Benfeng Xu}, \bibinfo{person}{Jin Xu}, \bibinfo{person}{An Yang}, \bibinfo{person}{Hao Yang},
  \bibinfo{person}{Jian Yang}, \bibinfo{person}{Jian Yang}, \bibinfo{person}{Shusheng Yang}, \bibinfo{person}{Yang Yao}, \bibinfo{person}{Bowen Yu}, \bibinfo{person}{Yu Bowen}, \bibinfo{person}{Hongyi Yuan}, \bibinfo{person}{Zheng Yuan}, \bibinfo{person}{Jianwei Zhang}, \bibinfo{person}{Xing Zhang}, \bibinfo{person}{Yichang Zhang}, \bibinfo{person}{Zhenru Zhang}, \bibinfo{person}{Chang Zhou}, \bibinfo{person}{Jingren Zhou}, \bibinfo{person}{Xiaohuan Zhou}, {and} \bibinfo{person}{Tianhang Zhu}.} \bibinfo{year}{2023}\natexlab{a}.
\newblock \showarticletitle{Qwen Technical Report}.
\newblock \bibinfo{journal}{\emph{ArXiv}} (\bibinfo{year}{2023}).
\newblock
\urldef\tempurl%
\url{https://doi.org/10.48550/arXiv.2309.16609}
\showURL{%
\tempurl}


\bibitem[Bai et~al\mbox{.}(2023b)]%
        {Bai2023QwenVLAF}
\bibfield{author}{\bibinfo{person}{Jinze Bai}, \bibinfo{person}{Shuai Bai}, \bibinfo{person}{Shusheng Yang}, \bibinfo{person}{Shijie Wang}, \bibinfo{person}{Sinan Tan}, \bibinfo{person}{Peng Wang}, \bibinfo{person}{Junyang Lin}, \bibinfo{person}{Chang Zhou}, {and} \bibinfo{person}{Jingren Zhou}.} \bibinfo{year}{2023}\natexlab{b}.
\newblock \showarticletitle{Qwen-VL: A Frontier Large Vision-Language Model with Versatile Abilities}.
\newblock \bibinfo{journal}{\emph{ArXiv}} (\bibinfo{year}{2023}).
\newblock
\urldef\tempurl%
\url{https://doi.org/10.48550/arXiv.2308.12966}
\showURL{%
\tempurl}


\bibitem[Bai et~al\mbox{.}(2025)]%
        {bai2025qwen2}
\bibfield{author}{\bibinfo{person}{Shuai Bai}, \bibinfo{person}{Keqin Chen}, \bibinfo{person}{Xuejing Liu}, \bibinfo{person}{Jialin Wang}, \bibinfo{person}{Wenbin Ge}, \bibinfo{person}{Sibo Song}, \bibinfo{person}{Kai Dang}, \bibinfo{person}{Peng Wang}, \bibinfo{person}{Shijie Wang}, \bibinfo{person}{Jun Tang}, {et~al\mbox{.}}} \bibinfo{year}{2025}\natexlab{}.
\newblock \showarticletitle{Qwen2. 5-vl technical report}.
\newblock \bibinfo{journal}{\emph{arXiv preprint arXiv:2502.13923}} (\bibinfo{year}{2025}).
\newblock


\bibitem[Bonatti et~al\mbox{.}(2024)]%
        {bonatti2024windows}
\bibfield{author}{\bibinfo{person}{Rogerio Bonatti}, \bibinfo{person}{Dan Zhao}, \bibinfo{person}{Francesco Bonacci}, \bibinfo{person}{Dillon Dupont}, \bibinfo{person}{Sara Abdali}, \bibinfo{person}{Yinheng Li}, \bibinfo{person}{Yadong Lu}, \bibinfo{person}{Justin Wagle}, \bibinfo{person}{Kazuhito Koishida}, \bibinfo{person}{Arthur Bucker}, {et~al\mbox{.}}} \bibinfo{year}{2024}\natexlab{}.
\newblock \showarticletitle{Windows agent arena: Evaluating multi-modal os agents at scale}.
\newblock \bibinfo{journal}{\emph{arXiv preprint arXiv:2409.08264}} (\bibinfo{year}{2024}).
\newblock


\bibitem[Cheng et~al\mbox{.}(2024)]%
        {cheng2024seeclick}
\bibfield{author}{\bibinfo{person}{Kanzhi Cheng}, \bibinfo{person}{Qiushi Sun}, \bibinfo{person}{Yougang Chu}, \bibinfo{person}{Fangzhi Xu}, \bibinfo{person}{Yantao Li}, \bibinfo{person}{Jianbing Zhang}, {and} \bibinfo{person}{Zhiyong Wu}.} \bibinfo{year}{2024}\natexlab{}.
\newblock \showarticletitle{Seeclick: Harnessing gui grounding for advanced visual gui agents}.
\newblock \bibinfo{journal}{\emph{arXiv preprint arXiv:2401.10935}} (\bibinfo{year}{2024}).
\newblock


\bibitem[DeepMind(2024)]%
        {googldeepmind2024}
\bibfield{author}{\bibinfo{person}{Google DeepMind}.} \bibinfo{year}{2024}\natexlab{}.
\newblock \bibinfo{title}{Gemini-2.0 (Project Mariner)}.
\newblock \bibinfo{howpublished}{\url{https://deepmind.google/technologies/project-mariner}}.
\newblock
\newblock
\shownote{Accessed: 2025-04-12}.


\bibitem[Dosovitskiy et~al\mbox{.}(2021)]%
        {Dosovitskiy2020AnII}
\bibfield{author}{\bibinfo{person}{Alexey Dosovitskiy}, \bibinfo{person}{Lucas Beyer}, \bibinfo{person}{Alexander Kolesnikov}, \bibinfo{person}{Dirk Weissenborn}, \bibinfo{person}{Xiaohua Zhai}, \bibinfo{person}{Thomas Unterthiner}, \bibinfo{person}{Mostafa Dehghani}, \bibinfo{person}{Matthias Minderer}, \bibinfo{person}{Georg Heigold}, \bibinfo{person}{Sylvain Gelly}, \bibinfo{person}{Jakob Uszkoreit}, {and} \bibinfo{person}{Neil Houlsby}.} \bibinfo{year}{2021}\natexlab{}.
\newblock \showarticletitle{An Image is Worth 16x16 Words: Transformers for Image Recognition at Scale}. In \bibinfo{booktitle}{\emph{9th International Conference on Learning Representations, {ICLR} 2021, Virtual Event, Austria, May 3-7, 2021}}. \bibinfo{publisher}{OpenReview.net}.
\newblock
\urldef\tempurl%
\url{https://openreview.net/forum?id=YicbFdNTTy}
\showURL{%
\tempurl}


\bibitem[Floridi and Chiriatti(2020)]%
        {floridi2020gpt}
\bibfield{author}{\bibinfo{person}{Luciano Floridi} {and} \bibinfo{person}{Massimo Chiriatti}.} \bibinfo{year}{2020}\natexlab{}.
\newblock \showarticletitle{GPT-3: Its nature, scope, limits, and consequences}.
\newblock \bibinfo{journal}{\emph{Minds and Machines}}  \bibinfo{volume}{30} (\bibinfo{year}{2020}), \bibinfo{pages}{681--694}.
\newblock


\bibitem[Gou et~al\mbox{.}(2024)]%
        {gou2024navigating}
\bibfield{author}{\bibinfo{person}{Boyu Gou}, \bibinfo{person}{Ruohan Wang}, \bibinfo{person}{Boyuan Zheng}, \bibinfo{person}{Yanan Xie}, \bibinfo{person}{Cheng Chang}, \bibinfo{person}{Yiheng Shu}, \bibinfo{person}{Huan Sun}, {and} \bibinfo{person}{Yu Su}.} \bibinfo{year}{2024}\natexlab{}.
\newblock \showarticletitle{Navigating the digital world as humans do: Universal visual grounding for gui agents}.
\newblock \bibinfo{journal}{\emph{arXiv preprint arXiv:2410.05243}} (\bibinfo{year}{2024}).
\newblock


\bibitem[Hong et~al\mbox{.}(2024)]%
        {hong2024cogagent}
\bibfield{author}{\bibinfo{person}{Wenyi Hong}, \bibinfo{person}{Weihan Wang}, \bibinfo{person}{Qingsong Lv}, \bibinfo{person}{Jiazheng Xu}, \bibinfo{person}{Wenmeng Yu}, \bibinfo{person}{Junhui Ji}, \bibinfo{person}{Yan Wang}, \bibinfo{person}{Zihan Wang}, \bibinfo{person}{Yuxiao Dong}, \bibinfo{person}{Ming Ding}, {et~al\mbox{.}}} \bibinfo{year}{2024}\natexlab{}.
\newblock \showarticletitle{Cogagent: A visual language model for gui agents}. In \bibinfo{booktitle}{\emph{Proceedings of the IEEE/CVF Conference on Computer Vision and Pattern Recognition}}. \bibinfo{pages}{14281--14290}.
\newblock


\bibitem[Hu et~al\mbox{.}(2024a)]%
        {202412.2294}
\bibfield{author}{\bibinfo{person}{Xueyu Hu}, \bibinfo{person}{Tao Xiong}, \bibinfo{person}{Biao Yi}, \bibinfo{person}{Zishu Wei}, \bibinfo{person}{Ruixuan Xiao}, \bibinfo{person}{Yurun Chen}, \bibinfo{person}{Jiasheng Ye}, \bibinfo{person}{Meiling Tao}, \bibinfo{person}{Xiangxin Zhou}, \bibinfo{person}{Ziyu Zhao}, \bibinfo{person}{Yuhuai Li}, \bibinfo{person}{Shengze Xu}, \bibinfo{person}{Shawn Wang}, \bibinfo{person}{Xinchen Xu}, \bibinfo{person}{Shuofei Qiao}, \bibinfo{person}{Kun Kuang}, \bibinfo{person}{Tieyong Zeng}, \bibinfo{person}{Liang Wang}, \bibinfo{person}{Jiwei Li}, \bibinfo{person}{Yuchen~Eleanor Jiang}, \bibinfo{person}{Wangchunshu Zhou}, \bibinfo{person}{Guoyin Wang}, \bibinfo{person}{Keting Yin}, \bibinfo{person}{Zhou Zhao}, \bibinfo{person}{Hongxia Yang}, \bibinfo{person}{Fan Wu}, \bibinfo{person}{Shengyu Zhang}, {and} \bibinfo{person}{Fei Wu}.} \bibinfo{year}{2024}\natexlab{a}.
\newblock \showarticletitle{OS Agents: A Survey on MLLM-Based Agents for General Computing Devices Use}.
\newblock \bibinfo{journal}{\emph{Preprints}} (\bibinfo{date}{December} \bibinfo{year}{2024}).
\newblock
\href{https://doi.org/10.20944/preprints202412.2294.v1}{doi:\nolinkurl{10.20944/preprints202412.2294.v1}}


\bibitem[Hu et~al\mbox{.}(2024b)]%
        {hu2024infiagentdabench}
\bibfield{author}{\bibinfo{person}{Xueyu Hu}, \bibinfo{person}{Ziyu Zhao}, \bibinfo{person}{Shuang Wei}, \bibinfo{person}{Ziwei Chai}, \bibinfo{person}{Qianli Ma}, \bibinfo{person}{Guoyin Wang}, \bibinfo{person}{Xuwu Wang}, \bibinfo{person}{Jing Su}, \bibinfo{person}{Jingjing Xu}, \bibinfo{person}{Ming Zhu}, \bibinfo{person}{Yao Cheng}, \bibinfo{person}{Jianbo Yuan}, \bibinfo{person}{Jiwei Li}, \bibinfo{person}{Kun Kuang}, \bibinfo{person}{Yang Yang}, \bibinfo{person}{Hongxia Yang}, {and} \bibinfo{person}{Fei Wu}.} \bibinfo{year}{2024}\natexlab{b}.
\newblock \showarticletitle{InfiAgent-DABench: Evaluating Agents on Data Analysis Tasks}.
\newblock \bibinfo{journal}{\emph{arXiv preprint arXiv:2401.05507}} (\bibinfo{year}{2024}).
\newblock


\bibitem[Huang et~al\mbox{.}(2024)]%
        {huang2024understanding}
\bibfield{author}{\bibinfo{person}{Xu Huang}, \bibinfo{person}{Weiwen Liu}, \bibinfo{person}{Xiaolong Chen}, \bibinfo{person}{Xingmei Wang}, \bibinfo{person}{Hao Wang}, \bibinfo{person}{Defu Lian}, \bibinfo{person}{Yasheng Wang}, \bibinfo{person}{Ruiming Tang}, {and} \bibinfo{person}{Enhong Chen}.} \bibinfo{year}{2024}\natexlab{}.
\newblock \showarticletitle{Understanding the planning of LLM agents: A survey}.
\newblock \bibinfo{journal}{\emph{arXiv preprint arXiv:2402.02716}} (\bibinfo{year}{2024}).
\newblock


\bibitem[Hurst et~al\mbox{.}(2024)]%
        {hurst2024gpt}
\bibfield{author}{\bibinfo{person}{Aaron Hurst}, \bibinfo{person}{Adam Lerer}, \bibinfo{person}{Adam~P Goucher}, \bibinfo{person}{Adam Perelman}, \bibinfo{person}{Aditya Ramesh}, \bibinfo{person}{Aidan Clark}, \bibinfo{person}{AJ Ostrow}, \bibinfo{person}{Akila Welihinda}, \bibinfo{person}{Alan Hayes}, \bibinfo{person}{Alec Radford}, {et~al\mbox{.}}} \bibinfo{year}{2024}\natexlab{}.
\newblock \showarticletitle{Gpt-4o system card}.
\newblock \bibinfo{journal}{\emph{arXiv preprint arXiv:2410.21276}} (\bibinfo{year}{2024}).
\newblock


\bibitem[Jiang et~al\mbox{.}(2023)]%
        {jiang2023iluvui}
\bibfield{author}{\bibinfo{person}{Yue Jiang}, \bibinfo{person}{Eldon Schoop}, \bibinfo{person}{Amanda Swearngin}, {and} \bibinfo{person}{Jeffrey Nichols}.} \bibinfo{year}{2023}\natexlab{}.
\newblock \showarticletitle{ILuvUI: Instruction-tuned LangUage-Vision modeling of UIs from Machine Conversations}.
\newblock \bibinfo{journal}{\emph{arXiv preprint arXiv:2310.04869}} (\bibinfo{year}{2023}).
\newblock


\bibitem[Jurmu et~al\mbox{.}(2008)]%
        {jurmu2008screenspot}
\bibfield{author}{\bibinfo{person}{Marko Jurmu}, \bibinfo{person}{Sebastian Boring}, {and} \bibinfo{person}{Jukka Riekki}.} \bibinfo{year}{2008}\natexlab{}.
\newblock \showarticletitle{ScreenSpot: Multidimensional resource discovery for distributed applications in smart spaces}. In \bibinfo{booktitle}{\emph{Proceedings of the 5th Annual International Conference on Mobile and Ubiquitous Systems: Computing, Networking, and Services}}. \bibinfo{pages}{1--9}.
\newblock


\bibitem[Li et~al\mbox{.}(2024c)]%
        {li2024llava}
\bibfield{author}{\bibinfo{person}{Bo Li}, \bibinfo{person}{Yuanhan Zhang}, \bibinfo{person}{Dong Guo}, \bibinfo{person}{Renrui Zhang}, \bibinfo{person}{Feng Li}, \bibinfo{person}{Hao Zhang}, \bibinfo{person}{Kaichen Zhang}, \bibinfo{person}{Peiyuan Zhang}, \bibinfo{person}{Yanwei Li}, \bibinfo{person}{Ziwei Liu}, {et~al\mbox{.}}} \bibinfo{year}{2024}\natexlab{c}.
\newblock \showarticletitle{Llava-onevision: Easy visual task transfer}.
\newblock \bibinfo{journal}{\emph{arXiv preprint arXiv:2408.03326}} (\bibinfo{year}{2024}).
\newblock


\bibitem[Li et~al\mbox{.}(2023)]%
        {li2023blip}
\bibfield{author}{\bibinfo{person}{Junnan Li}, \bibinfo{person}{Dongxu Li}, \bibinfo{person}{Silvio Savarese}, {and} \bibinfo{person}{Steven Hoi}.} \bibinfo{year}{2023}\natexlab{}.
\newblock \showarticletitle{Blip-2: Bootstrapping language-image pre-training with frozen image encoders and large language models}. In \bibinfo{booktitle}{\emph{International conference on machine learning}}. PMLR, \bibinfo{pages}{19730--19742}.
\newblock


\bibitem[Li et~al\mbox{.}(2025)]%
        {li2025screenspot}
\bibfield{author}{\bibinfo{person}{Kaixin Li}, \bibinfo{person}{Hongzhan Lin}, \bibinfo{person}{Ziyang Luo}, \bibinfo{person}{Yuchen Tian}, \bibinfo{person}{Jing Ma}, \bibinfo{person}{Zhiyong Huang}, \bibinfo{person}{Tat-Seng Chua}, {et~al\mbox{.}}} \bibinfo{year}{2025}\natexlab{}.
\newblock \showarticletitle{Screenspot-pro: Gui grounding for professional high-resolution computer use}. In \bibinfo{booktitle}{\emph{Workshop on Reasoning and Planning for Large Language Models}}.
\newblock


\bibitem[Li et~al\mbox{.}(2024b)]%
        {li2024inficodereval}
\bibfield{author}{\bibinfo{person}{Linyi Li}, \bibinfo{person}{Shijie Geng}, \bibinfo{person}{Zhenwen Li}, \bibinfo{person}{Yibo He}, \bibinfo{person}{Hao Yu}, \bibinfo{person}{Ziyue Hua}, \bibinfo{person}{Guanghan Ning}, \bibinfo{person}{Siwei Wang}, \bibinfo{person}{Tao Xie}, {and} \bibinfo{person}{Hongxia Yang}.} \bibinfo{year}{2024}\natexlab{b}.
\newblock \showarticletitle{InfiBench: Evaluating the Question-Answering Capabilities of Code Large Language Models}.
\newblock \bibinfo{journal}{\emph{arXiv preprint arXiv:2404.07940}} (\bibinfo{year}{2024}).
\newblock


\bibitem[Li et~al\mbox{.}(2024a)]%
        {li2024effects}
\bibfield{author}{\bibinfo{person}{Wei Li}, \bibinfo{person}{William Bishop}, \bibinfo{person}{Alice Li}, \bibinfo{person}{Chris Rawles}, \bibinfo{person}{Folawiyo Campbell-Ajala}, \bibinfo{person}{Divya Tyamagundlu}, {and} \bibinfo{person}{Oriana Riva}.} \bibinfo{year}{2024}\natexlab{a}.
\newblock \showarticletitle{On the effects of data scale on computer control agents}.
\newblock \bibinfo{journal}{\emph{arXiv e-prints}} (\bibinfo{year}{2024}), \bibinfo{pages}{arXiv--2406}.
\newblock


\bibitem[Li et~al\mbox{.}(2020)]%
        {li2020widget}
\bibfield{author}{\bibinfo{person}{Yang Li}, \bibinfo{person}{Luheng Li, Gangaand~He}, \bibinfo{person}{Jingjie Zheng}, \bibinfo{person}{Hong Li}, {and} \bibinfo{person}{Zhiwei Guan}.} \bibinfo{year}{2020}\natexlab{}.
\newblock \showarticletitle{Widget Captioning: Generating Natural Language Description for Mobile User Interface Elements}.
\newblock \bibinfo{journal}{\emph{arXiv preprint arXiv:2010.04295}} (\bibinfo{year}{2020}).
\newblock


\bibitem[Liang et~al\mbox{.}(2024)]%
        {liang2024survey}
\bibfield{author}{\bibinfo{person}{Zijing Liang}, \bibinfo{person}{Yanjie Xu}, \bibinfo{person}{Yifan Hong}, \bibinfo{person}{Penghui Shang}, \bibinfo{person}{Qi Wang}, \bibinfo{person}{Qiang Fu}, {and} \bibinfo{person}{Ke Liu}.} \bibinfo{year}{2024}\natexlab{}.
\newblock \showarticletitle{A Survey of Multimodel Large Language Models}. In \bibinfo{booktitle}{\emph{Proceedings of the 3rd International Conference on Computer, Artificial Intelligence and Control Engineering}}. \bibinfo{pages}{405--409}.
\newblock


\bibitem[Lin et~al\mbox{.}(2024)]%
        {lin2024showui}
\bibfield{author}{\bibinfo{person}{Kevin~Qinghong Lin}, \bibinfo{person}{Linjie Li}, \bibinfo{person}{Difei Gao}, \bibinfo{person}{Zhengyuan Yang}, \bibinfo{person}{Zechen Bai}, \bibinfo{person}{Weixian Lei}, \bibinfo{person}{Lijuan Wang}, {and} \bibinfo{person}{Mike~Zheng Shou}.} \bibinfo{year}{2024}\natexlab{}.
\newblock \showarticletitle{Showui: One vision-language-action model for generalist gui agent}. In \bibinfo{booktitle}{\emph{NeurIPS 2024 Workshop on Open-World Agents}}.
\newblock


\bibitem[Lin et~al\mbox{.}(2014)]%
        {lin2014microsoft}
\bibfield{author}{\bibinfo{person}{Tsung-Yi Lin}, \bibinfo{person}{Michael Maire}, \bibinfo{person}{Serge Belongie}, \bibinfo{person}{James Hays}, \bibinfo{person}{Pietro Perona}, \bibinfo{person}{Deva Ramanan}, \bibinfo{person}{Piotr Doll{\'a}r}, {and} \bibinfo{person}{C~Lawrence Zitnick}.} \bibinfo{year}{2014}\natexlab{}.
\newblock \showarticletitle{Microsoft coco: Common objects in context}. In \bibinfo{booktitle}{\emph{Computer vision--ECCV 2014: 13th European conference, zurich, Switzerland, September 6-12, 2014, proceedings, part v 13}}. Springer, \bibinfo{pages}{740--755}.
\newblock


\bibitem[Liu et~al\mbox{.}(2023)]%
        {Liu2023VisualIT}
\bibfield{author}{\bibinfo{person}{Haotian Liu}, \bibinfo{person}{Chunyuan Li}, \bibinfo{person}{Qingyang Wu}, {and} \bibinfo{person}{Yong~Jae Lee}.} \bibinfo{year}{2023}\natexlab{}.
\newblock \showarticletitle{Visual Instruction Tuning}. In \bibinfo{booktitle}{\emph{Advances in Neural Information Processing Systems 36: Annual Conference on Neural Information Processing Systems 2023, NeurIPS 2023, New Orleans, LA, USA, December 10 - 16, 2023}}, \bibfield{editor}{\bibinfo{person}{Alice Oh}, \bibinfo{person}{Tristan Naumann}, \bibinfo{person}{Amir Globerson}, \bibinfo{person}{Kate Saenko}, \bibinfo{person}{Moritz Hardt}, {and} \bibinfo{person}{Sergey Levine}} (Eds.).
\newblock
\urldef\tempurl%
\url{http://papers.nips.cc/paper\_files/paper/2023/hash/6dcf277ea32ce3288914faf369fe6de0-Abstract-Conference.html}
\showURL{%
\tempurl}


\bibitem[Liu et~al\mbox{.}(2024)]%
        {liu2024infimm}
\bibfield{author}{\bibinfo{person}{Haogeng Liu}, \bibinfo{person}{Quanzeng You}, \bibinfo{person}{Yiqi Wang}, \bibinfo{person}{Xiaotian Han}, \bibinfo{person}{Bohan Zhai}, \bibinfo{person}{Yongfei Liu}, \bibinfo{person}{Wentao Chen}, \bibinfo{person}{Yiren Jian}, \bibinfo{person}{Yunzhe Tao}, \bibinfo{person}{Jianbo Yuan}, \bibinfo{person}{Ran He}, {and} \bibinfo{person}{Hongxia Yang}.} \bibinfo{year}{2024}\natexlab{}.
\newblock \showarticletitle{InfiMM: Advancing Multimodal Understanding with an Open-Sourced Visual Language Model}. In \bibinfo{booktitle}{\emph{Annual Meeting of the Association for Computational Linguistics}}.
\newblock


\bibitem[Liu et~al\mbox{.}(2025)]%
        {liu2025infiguiagent}
\bibfield{author}{\bibinfo{person}{Yuhang Liu}, \bibinfo{person}{Pengxiang Li}, \bibinfo{person}{Zishu Wei}, \bibinfo{person}{Congkai Xie}, \bibinfo{person}{Xueyu Hu}, \bibinfo{person}{Xinchen Xu}, \bibinfo{person}{Shengyu Zhang}, \bibinfo{person}{Xiaotian Han}, \bibinfo{person}{Hongxia Yang}, {and} \bibinfo{person}{Fei Wu}.} \bibinfo{year}{2025}\natexlab{}.
\newblock \showarticletitle{InfiGUIAgent: A Multimodal Generalist GUI Agent with Native Reasoning and Reflection}.
\newblock \bibinfo{journal}{\emph{arXiv preprint arXiv:2501.04575}} (\bibinfo{year}{2025}).
\newblock


\bibitem[Liu et~al\mbox{.}(2022)]%
        {liu2022convnet}
\bibfield{author}{\bibinfo{person}{Zhuang Liu}, \bibinfo{person}{Hanzi Mao}, \bibinfo{person}{Chao-Yuan Wu}, \bibinfo{person}{Christoph Feichtenhofer}, \bibinfo{person}{Trevor Darrell}, {and} \bibinfo{person}{Saining Xie}.} \bibinfo{year}{2022}\natexlab{}.
\newblock \showarticletitle{A convnet for the 2020s}. In \bibinfo{booktitle}{\emph{Proceedings of the IEEE/CVF conference on computer vision and pattern recognition}}. \bibinfo{pages}{11976--11986}.
\newblock


\bibitem[Lu et~al\mbox{.}(2025)]%
        {lu2025ui}
\bibfield{author}{\bibinfo{person}{Zhengxi Lu}, \bibinfo{person}{Yuxiang Chai}, \bibinfo{person}{Yaxuan Guo}, \bibinfo{person}{Xi Yin}, \bibinfo{person}{Liang Liu}, \bibinfo{person}{Hao Wang}, \bibinfo{person}{Guanjing Xiong}, {and} \bibinfo{person}{Hongsheng Li}.} \bibinfo{year}{2025}\natexlab{}.
\newblock \showarticletitle{UI-R1: Enhancing Action Prediction of GUI Agents by Reinforcement Learning}.
\newblock \bibinfo{journal}{\emph{arXiv preprint arXiv:2503.21620}} (\bibinfo{year}{2025}).
\newblock


\bibitem[OpenAI(2023)]%
        {openai2023gpt4v}
\bibfield{author}{\bibinfo{person}{OpenAI}.} \bibinfo{year}{2023}\natexlab{}.
\newblock \bibinfo{title}{GPT-4V(ision) System Card}.
\newblock
\urldef\tempurl%
\url{https://cdn.openai.com/papers/GPTV_System_Card.pdf}
\showURL{%
\tempurl}


\bibitem[OpenAI(2024)]%
        {gpt4o}
\bibfield{author}{\bibinfo{person}{OpenAI}.} \bibinfo{year}{2024}\natexlab{}.
\newblock \bibinfo{title}{GPT-4o}.
\newblock
\urldef\tempurl%
\url{https://openai.com/index/hello-gpt-4o/}
\showURL{%
\tempurl}
\newblock
\shownote{Accessed: 2025-01-03}.


\bibitem[Pan et~al\mbox{.}(2024)]%
        {pan2024chain}
\bibfield{author}{\bibinfo{person}{Zhenyu Pan}, \bibinfo{person}{Haozheng Luo}, \bibinfo{person}{Manling Li}, {and} \bibinfo{person}{Han Liu}.} \bibinfo{year}{2024}\natexlab{}.
\newblock \showarticletitle{Chain-of-action: Faithful and multimodal question answering through large language models}.
\newblock \bibinfo{journal}{\emph{arXiv preprint arXiv:2403.17359}} (\bibinfo{year}{2024}).
\newblock


\bibitem[Peng et~al\mbox{.}(2023)]%
        {peng2023kosmos}
\bibfield{author}{\bibinfo{person}{Zhiliang Peng}, \bibinfo{person}{Wenhui Wang}, \bibinfo{person}{Li Dong}, \bibinfo{person}{Yaru Hao}, \bibinfo{person}{Shaohan Huang}, \bibinfo{person}{Shuming Ma}, {and} \bibinfo{person}{Furu Wei}.} \bibinfo{year}{2023}\natexlab{}.
\newblock \showarticletitle{Kosmos-2: Grounding multimodal large language models to the world}.
\newblock \bibinfo{journal}{\emph{arXiv preprint arXiv:2306.14824}} (\bibinfo{year}{2023}).
\newblock


\bibitem[Qin et~al\mbox{.}(2025)]%
        {qin2025ui}
\bibfield{author}{\bibinfo{person}{Yujia Qin}, \bibinfo{person}{Yining Ye}, \bibinfo{person}{Junjie Fang}, \bibinfo{person}{Haoming Wang}, \bibinfo{person}{Shihao Liang}, \bibinfo{person}{Shizuo Tian}, \bibinfo{person}{Junda Zhang}, \bibinfo{person}{Jiahao Li}, \bibinfo{person}{Yunxin Li}, \bibinfo{person}{Shijue Huang}, {et~al\mbox{.}}} \bibinfo{year}{2025}\natexlab{}.
\newblock \showarticletitle{UI-TARS: Pioneering Automated GUI Interaction with Native Agents}.
\newblock \bibinfo{journal}{\emph{arXiv preprint arXiv:2501.12326}} (\bibinfo{year}{2025}).
\newblock


\bibitem[Radford et~al\mbox{.}(2021)]%
        {radford2021learning}
\bibfield{author}{\bibinfo{person}{Alec Radford}, \bibinfo{person}{Jong~Wook Kim}, \bibinfo{person}{Chris Hallacy}, \bibinfo{person}{Aditya Ramesh}, \bibinfo{person}{Gabriel Goh}, \bibinfo{person}{Sandhini Agarwal}, \bibinfo{person}{Girish Sastry}, \bibinfo{person}{Amanda Askell}, \bibinfo{person}{Pamela Mishkin}, \bibinfo{person}{Jack Clark}, {et~al\mbox{.}}} \bibinfo{year}{2021}\natexlab{}.
\newblock \showarticletitle{Learning transferable visual models from natural language supervision}. In \bibinfo{booktitle}{\emph{International conference on machine learning}}. PMLR, \bibinfo{pages}{8748--8763}.
\newblock


\bibitem[Rawles et~al\mbox{.}(2024)]%
        {rawles2024androidworld}
\bibfield{author}{\bibinfo{person}{Christopher Rawles}, \bibinfo{person}{Sarah Clinckemaillie}, \bibinfo{person}{Yifan Chang}, \bibinfo{person}{Jonathan Waltz}, \bibinfo{person}{Gabrielle Lau}, \bibinfo{person}{Marybeth Fair}, \bibinfo{person}{Alice Li}, \bibinfo{person}{William Bishop}, \bibinfo{person}{Wei Li}, \bibinfo{person}{Folawiyo Campbell-Ajala}, {et~al\mbox{.}}} \bibinfo{year}{2024}\natexlab{}.
\newblock \showarticletitle{Androidworld: A dynamic benchmarking environment for autonomous agents}.
\newblock \bibinfo{journal}{\emph{arXiv preprint arXiv:2405.14573}} (\bibinfo{year}{2024}).
\newblock


\bibitem[Sheng et~al\mbox{.}(2024)]%
        {sheng2024hybridflow}
\bibfield{author}{\bibinfo{person}{Guangming Sheng}, \bibinfo{person}{Chi Zhang}, \bibinfo{person}{Zilingfeng Ye}, \bibinfo{person}{Xibin Wu}, \bibinfo{person}{Wang Zhang}, \bibinfo{person}{Ru Zhang}, \bibinfo{person}{Yanghua Peng}, \bibinfo{person}{Haibin Lin}, {and} \bibinfo{person}{Chuan Wu}.} \bibinfo{year}{2024}\natexlab{}.
\newblock \showarticletitle{HybridFlow: A Flexible and Efficient RLHF Framework}.
\newblock \bibinfo{journal}{\emph{arXiv preprint arXiv: 2409.19256}} (\bibinfo{year}{2024}).
\newblock


\bibitem[Sun et~al\mbox{.}(2025)]%
        {sun2025gui}
\bibfield{author}{\bibinfo{person}{Yuchen Sun}, \bibinfo{person}{Shanhui Zhao}, \bibinfo{person}{Tao Yu}, \bibinfo{person}{Hao Wen}, \bibinfo{person}{Samith Va}, \bibinfo{person}{Mengwei Xu}, \bibinfo{person}{Yuanchun Li}, {and} \bibinfo{person}{Chongyang Zhang}.} \bibinfo{year}{2025}\natexlab{}.
\newblock \showarticletitle{GUI-Xplore: Empowering Generalizable GUI Agents with One Exploration}.
\newblock \bibinfo{journal}{\emph{arXiv preprint arXiv:2503.17709}} (\bibinfo{year}{2025}).
\newblock


\bibitem[Team et~al\mbox{.}(2025)]%
        {team2025kimi}
\bibfield{author}{\bibinfo{person}{Kimi Team}, \bibinfo{person}{Angang Du}, \bibinfo{person}{Bohong Yin}, \bibinfo{person}{Bowei Xing}, \bibinfo{person}{Bowen Qu}, \bibinfo{person}{Bowen Wang}, \bibinfo{person}{Cheng Chen}, \bibinfo{person}{Chenlin Zhang}, \bibinfo{person}{Chenzhuang Du}, \bibinfo{person}{Chu Wei}, {et~al\mbox{.}}} \bibinfo{year}{2025}\natexlab{}.
\newblock \showarticletitle{Kimi-VL Technical Report}.
\newblock \bibinfo{journal}{\emph{arXiv preprint arXiv:2504.07491}} (\bibinfo{year}{2025}).
\newblock


\bibitem[Team(2025)]%
        {qwq32b}
\bibfield{author}{\bibinfo{person}{Qwen Team}.} \bibinfo{year}{2025}\natexlab{}.
\newblock \bibinfo{title}{QwQ-32B: Embracing the Power of Reinforcement Learning}.
\newblock
\urldef\tempurl%
\url{https://qwenlm.github.io/blog/qwq-32b/}
\showURL{%
\tempurl}


\bibitem[Touvron et~al\mbox{.}(2023)]%
        {Touvron2023LLaMAOA}
\bibfield{author}{\bibinfo{person}{Hugo Touvron}, \bibinfo{person}{Thibaut Lavril}, \bibinfo{person}{Gautier Izacard}, \bibinfo{person}{Xavier Martinet}, \bibinfo{person}{Marie-Anne Lachaux}, \bibinfo{person}{Timoth{\'e}e Lacroix}, \bibinfo{person}{Baptiste Rozi{\`e}re}, \bibinfo{person}{Naman Goyal}, \bibinfo{person}{Eric Hambro}, \bibinfo{person}{Faisal Azhar}, \bibinfo{person}{Aurelien Rodriguez}, \bibinfo{person}{Armand Joulin}, \bibinfo{person}{Edouard Grave}, {and} \bibinfo{person}{Guillaume Lample}.} \bibinfo{year}{2023}\natexlab{}.
\newblock \showarticletitle{LLaMA: Open and Efficient Foundation Language Models}.
\newblock \bibinfo{journal}{\emph{ArXiv}} (\bibinfo{year}{2023}).
\newblock
\urldef\tempurl%
\url{https://doi.org/10.48550/arXiv.2302.13971}
\showURL{%
\tempurl}


\bibitem[Wang et~al\mbox{.}(2024)]%
        {wang2024qwen2}
\bibfield{author}{\bibinfo{person}{Peng Wang}, \bibinfo{person}{Shuai Bai}, \bibinfo{person}{Sinan Tan}, \bibinfo{person}{Shijie Wang}, \bibinfo{person}{Zhihao Fan}, \bibinfo{person}{Jinze Bai}, \bibinfo{person}{Keqin Chen}, \bibinfo{person}{Xuejing Liu}, \bibinfo{person}{Jialin Wang}, \bibinfo{person}{Wenbin Ge}, {et~al\mbox{.}}} \bibinfo{year}{2024}\natexlab{}.
\newblock \showarticletitle{Qwen2-vl: Enhancing vision-language model's perception of the world at any resolution}.
\newblock \bibinfo{journal}{\emph{arXiv preprint arXiv:2409.12191}} (\bibinfo{year}{2024}).
\newblock


\bibitem[Wang et~al\mbox{.}(2023)]%
        {wang2023cogvlm}
\bibfield{author}{\bibinfo{person}{Weihan Wang}, \bibinfo{person}{Qingsong Lv}, \bibinfo{person}{Wenmeng Yu}, \bibinfo{person}{Wenyi Hong}, \bibinfo{person}{Ji Qi}, \bibinfo{person}{Yan Wang}, \bibinfo{person}{Junhui Ji}, \bibinfo{person}{Zhuoyi Yang}, \bibinfo{person}{Lei Zhao}, \bibinfo{person}{Xixuan Song}, {et~al\mbox{.}}} \bibinfo{year}{2023}\natexlab{}.
\newblock \showarticletitle{Cogvlm: Visual expert for pretrained language models}.
\newblock \bibinfo{journal}{\emph{arXiv preprint arXiv:2311.03079}} (\bibinfo{year}{2023}).
\newblock


\bibitem[Wen et~al\mbox{.}(2023)]%
        {wen2023autodroid}
\bibfield{author}{\bibinfo{person}{Hao Wen}, \bibinfo{person}{Yuanchun Li}, \bibinfo{person}{Guohong Liu}, \bibinfo{person}{Shanhui Zhao}, \bibinfo{person}{Tao Yu}, \bibinfo{person}{Toby Jia-Jun Li}, \bibinfo{person}{Shiqi Jiang}, \bibinfo{person}{Yunhao Liu}, \bibinfo{person}{Yaqin Zhang}, {and} \bibinfo{person}{Yunxin Liu}.} \bibinfo{year}{2023}\natexlab{}.
\newblock \showarticletitle{AutoDroid: LLM-powered Task Automation in Android}.
\newblock \bibinfo{journal}{\emph{arXiv preprint arXiv:2308.15272}} (\bibinfo{year}{2023}).
\newblock


\bibitem[Wu et~al\mbox{.}(2024)]%
        {wu2024atlas}
\bibfield{author}{\bibinfo{person}{Zhiyong Wu}, \bibinfo{person}{Zhenyu Wu}, \bibinfo{person}{Fangzhi Xu}, \bibinfo{person}{Yian Wang}, \bibinfo{person}{Qiushi Sun}, \bibinfo{person}{Chengyou Jia}, \bibinfo{person}{Kanzhi Cheng}, \bibinfo{person}{Zichen Ding}, \bibinfo{person}{Liheng Chen}, \bibinfo{person}{Paul~Pu Liang}, {et~al\mbox{.}}} \bibinfo{year}{2024}\natexlab{}.
\newblock \showarticletitle{Os-atlas: A foundation action model for generalist gui agents}.
\newblock \bibinfo{journal}{\emph{arXiv preprint arXiv:2410.23218}} (\bibinfo{year}{2024}).
\newblock


\bibitem[Xia and Luo(2025)]%
        {xia2025gui}
\bibfield{author}{\bibinfo{person}{Xiaobo Xia} {and} \bibinfo{person}{Run Luo}.} \bibinfo{year}{2025}\natexlab{}.
\newblock \showarticletitle{GUI-R1: A Generalist R1-Style Vision-Language Action Model For GUI Agents}.
\newblock \bibinfo{journal}{\emph{arXiv preprint arXiv:2504.10458}} (\bibinfo{year}{2025}).
\newblock


\bibitem[Xiao et~al\mbox{.}(2021)]%
        {xiao2021lawformer}
\bibfield{author}{\bibinfo{person}{Chaojun Xiao}, \bibinfo{person}{Xueyu Hu}, \bibinfo{person}{Zhiyuan Liu}, \bibinfo{person}{Cunchao Tu}, {and} \bibinfo{person}{Maosong Sun}.} \bibinfo{year}{2021}\natexlab{}.
\newblock \showarticletitle{Lawformer: A pre-trained language model for chinese legal long documents}.
\newblock \bibinfo{journal}{\emph{AI Open}}  \bibinfo{volume}{2} (\bibinfo{year}{2021}), \bibinfo{pages}{79--84}.
\newblock


\bibitem[Yang et~al\mbox{.}(2023)]%
        {yang2023set}
\bibfield{author}{\bibinfo{person}{Jianwei Yang}, \bibinfo{person}{Hao Zhang}, \bibinfo{person}{Feng Li}, \bibinfo{person}{Xueyan Zou}, \bibinfo{person}{Chunyuan Li}, {and} \bibinfo{person}{Jianfeng Gao}.} \bibinfo{year}{2023}\natexlab{}.
\newblock \showarticletitle{Set-of-mark prompting unleashes extraordinary visual grounding in gpt-4v}.
\newblock \bibinfo{journal}{\emph{arXiv preprint arXiv:2310.11441}} (\bibinfo{year}{2023}).
\newblock


\bibitem[Yang et~al\mbox{.}(2024)]%
        {yang2024aria}
\bibfield{author}{\bibinfo{person}{Yuhao Yang}, \bibinfo{person}{Yue Wang}, \bibinfo{person}{Dongxu Li}, \bibinfo{person}{Ziyang Luo}, \bibinfo{person}{Bei Chen}, \bibinfo{person}{Chao Huang}, {and} \bibinfo{person}{Junnan Li}.} \bibinfo{year}{2024}\natexlab{}.
\newblock \showarticletitle{Aria-UI: Visual Grounding for GUI Instructions}.
\newblock \bibinfo{journal}{\emph{arXiv preprint arXiv:2412.16256}} (\bibinfo{year}{2024}).
\newblock


\bibitem[Yaowei~Zheng(2025)]%
        {zheng2025easyr1}
\bibfield{author}{\bibinfo{person}{Shenzhi Wang Zhangchi Feng Dongdong Kuang Yuwen~Xiong Yaowei~Zheng, Junting~Lu}.} \bibinfo{year}{2025}\natexlab{}.
\newblock \bibinfo{title}{EasyR1: An Efficient, Scalable, Multi-Modality RL Training Framework}.
\newblock \bibinfo{howpublished}{\url{https://github.com/hiyouga/EasyR1}}.
\newblock


\bibitem[You et~al\mbox{.}(2025)]%
        {you2025ferret}
\bibfield{author}{\bibinfo{person}{Keen You}, \bibinfo{person}{Haotian Zhang}, \bibinfo{person}{Eldon Schoop}, \bibinfo{person}{Floris Weers}, \bibinfo{person}{Amanda Swearngin}, \bibinfo{person}{Jeffrey Nichols}, \bibinfo{person}{Yinfei Yang}, {and} \bibinfo{person}{Zhe Gan}.} \bibinfo{year}{2025}\natexlab{}.
\newblock \showarticletitle{Ferret-ui: Grounded mobile ui understanding with multimodal llms}. In \bibinfo{booktitle}{\emph{European Conference on Computer Vision}}. Springer, \bibinfo{pages}{240--255}.
\newblock


\bibitem[Zhang et~al\mbox{.}(2023)]%
        {zhang2023appagent}
\bibfield{author}{\bibinfo{person}{Chi Zhang}, \bibinfo{person}{Zhao Yang}, \bibinfo{person}{Jiaxuan Liu}, \bibinfo{person}{Yucheng Han}, \bibinfo{person}{Xin Chen}, \bibinfo{person}{Zebiao Huang}, \bibinfo{person}{Bin Fu}, {and} \bibinfo{person}{Gang Yu}.} \bibinfo{year}{2023}\natexlab{}.
\newblock \showarticletitle{Appagent: Multimodal agents as smartphone users}.
\newblock \bibinfo{journal}{\emph{arXiv preprint arXiv:2312.13771}} (\bibinfo{year}{2023}).
\newblock


\end{thebibliography}

\appendix

\end{document}